\documentclass[sigconf,screen,nonacm]{acmart}
\AtBeginDocument{%
  }

\setcopyright{acmlicensed}
\copyrightyear{2025}
\acmYear{2025}
\acmDOI{Under Submission}
\acmConference[AISec’25]{the 2025 Workshop on Artificial Intelligence and Security, October 17, 2025, Taipei, Taiwan}{October 17,
  2025}{Taipei, Taiwan}
\acmISBN{978-1-4503-XXXX-X/2025/06}

\usepackage{colortbl}
\usepackage[cachedir=minted,frozencache=true]{minted}
\usepackage{xcolor}
\usepackage{booktabs}
\usepackage{pifont}
\usepackage{siunitx}
\usepackage{url}
\usepackage[most]{tcolorbox}
\usepackage{tabularx}     
\usepackage{enumitem} 
\newcommand{\Supported}{\textcolor{green!60!black}{\checkmark}}          
\newcommand{\Limited}{\textcolor{orange!85!black}{\(\blacktriangle\)}}   
\newcommand{\Unsupported}{\textcolor{red!70!black}{\(\times\)}}          

\usepackage{array}
\newcolumntype{L}[1]{>{\raggedright\arraybackslash}p{#1}} 
\newcolumntype{Y}{>{\raggedright\arraybackslash}X}         
\definecolor{grayrow}{gray}{0.95}
\begin{document}

\title{I Know Which LLM Wrote Your Code Last Summer: \\
LLM generated Code Stylometry for Authorship Attribution}

\author{Tamas Bisztray}
\orcid{0000-0003-2626-3434}
\affiliation{%
  \institution{University of Oslo}
  \city{Oslo}
  \country{Norway}
}
\email{tamasbi@ifi.uio.no}

\author{Bilel Cherif}
\orcid{0009-0006-0095-106X}
\affiliation{%
  \institution{Technology Innovation Institute}
  \city{Abu Dhabi}
  \country{United Arab Emirates}
}
\email{bilel.cherif@tii.ae}

\author{Richard A. Dubniczky}
\orcid{0009-0003-3951-1932}
\affiliation{%
  \institution{Eötvös Lóránd University}
  \city{Budapest}
  \country{Hungary}
}
\email{richard@dubniczky.com}

\author{Nils Gruschka}
\orcid{0000-0001-7360-8314}
\affiliation{%
  \institution{University of Oslo}
  \city{Oslo}
  \country{Norway}
}
\email{nilsgrus@ifi.uio.no}

\author{Bertalan Borsos}
\orcid{0009-0000-8718-8285}
\affiliation{%
  \institution{Eötvös Lóránd University}
  \city{Budapest}
  \country{Hungary}
}
\email{borsosb@inf.elte.hu}

\author{Mohamed Amine Ferrag}
\orcid{0000-0002-0632-3172}
\affiliation{%
  \institution{Guelma University}
  \city{Guelma}
  \country{Algeria}
}
\email{mohamed.amine.ferrag@gmail.com}

\author{Attila Kovacs}
\orcid{0000-0002-1858-7618}
\affiliation{%
  \institution{Eötvös Lóránd University}
  \city{Budapest}
  \country{Hungary}
}
\email{attila@inf.elte.hu}

\author{Vasileios Mavroeidis}
\orcid{0000-0003-1097-0599}
\affiliation{%
  \institution{University of Oslo, Cyentific AS}
  \city{Oslo}
  \country{Norway}
}
\email{vasileim@ifi.uio.no}

\author{Norbert Tihanyi}
\orcid{0000-0002-9002-5935}
\affiliation{%
  \institution{Technology Innovation Institute}
  \city{Abu Dhabi}
  \country{United Arab Emirates}
}
\email{norbert.tihanyi@tii.ae}
\authornote{Corresponding author}
\renewcommand{\shortauthors}{Bisztray et al.}

\begin{abstract}
Detecting AI-generated code,  deepfakes, and other synthetic content is an emerging research challenge. As code generated by Large Language Models (LLMs) becomes more common, identifying the specific model behind each sample is increasingly important. This paper presents the first systematic study of LLM authorship attribution for C programs. We released CodeT5-Authorship, a novel model that uses only the encoder layers from the original CodeT5 encoder-decoder architecture, discarding the decoder to focus on classification. Our model's encoder output (first token) is passed through a two-layer classification head with GELU activation and dropout, producing a probability distribution over possible authors. To evaluate our approach, we introduce LLM-AuthorBench, a benchmark of 32,000 compilable C programs generated by eight state-of-the-art LLMs across diverse tasks. We compare our model to seven traditional ML classifiers and eight fine-tuned transformer models, including BERT, RoBERTa, CodeBERT, ModernBERT, DistilBERT, DeBERTa-V3, Longformer, and LoRA-fine-tuned Qwen2-1.5B. In binary classification, our model achieves 97.56\% accuracy in distinguishing C programs generated by closely related models such as GPT-4.1 and GPT-4o, and 95.40\% accuracy for multi-class attribution among five leading LLMs (Gemini 2.5 Flash, Claude 3.5 Haiku, GPT-4.1, Llama 3.3, and DeepSeek-V3). To support open science, we release the CodeT5-Authorship architecture, the LLM-AuthorBench benchmark, and all relevant Google Colab scripts on GitHub: \url{https://github.com/LLMauthorbench/}.

\end{abstract}



\received{*}
\received[revised]{*}
\received[accepted]{*}
\maketitle
\section{Introduction}

\begin{figure}[h]
    \centering
    \includegraphics[width=0.9\linewidth]{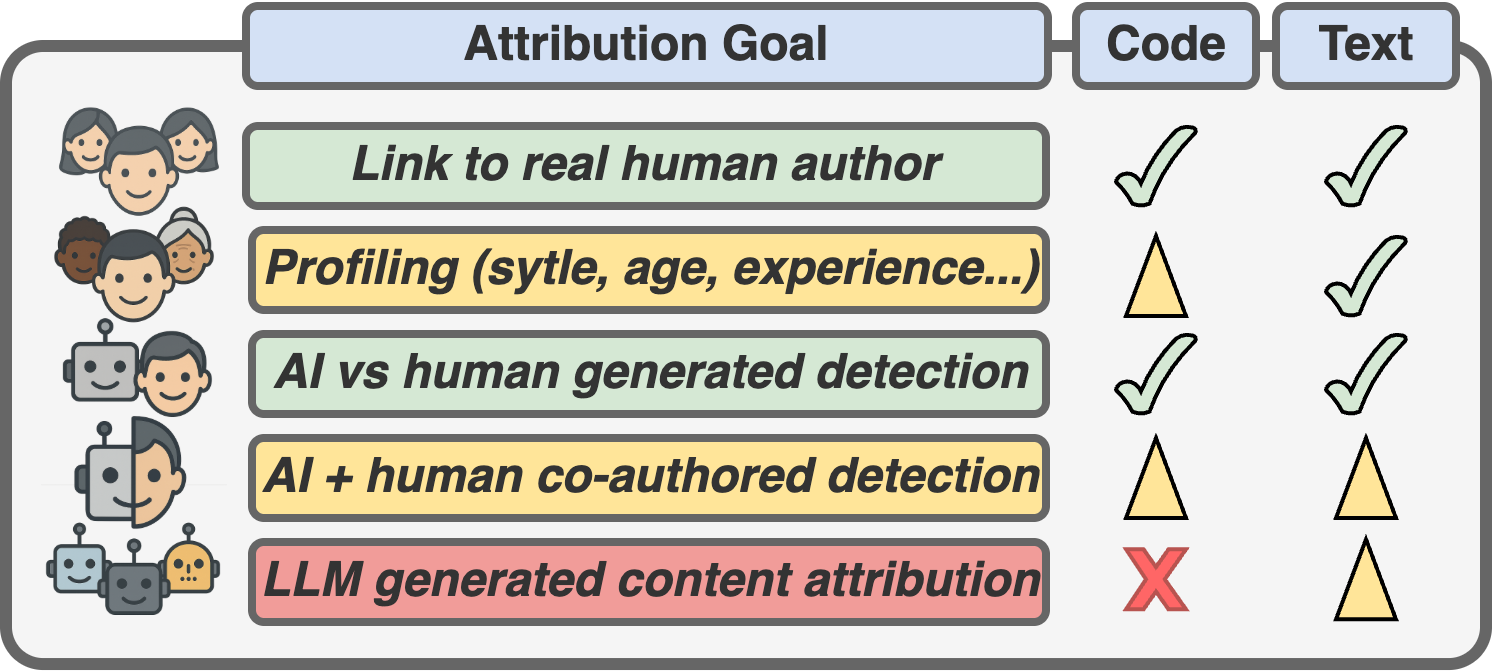}
    \caption{Attribution goals for source code and text.}
    \textbf{Legend:} \Supported{} well-researched; \Limited{} limited research; \Unsupported{} not investigated
    \label{fig:enter-label}
\end{figure}

As AI-generated content is getting more widespread, authorship attribution---the task of linking unknown content to its creator---is becoming essential for accountability in areas ranging from preventing plagiarism to ensuring legal integrity~\cite{he_authorship_2024,boenninghoff_explainable_2019,kumarage_survey_2024}. With recent advances in \textit{Large Language Models} (LLMs), this challenge has become even more significant, as LLMs can now automatically generate high-quality text and code. Authorship analysis comprises five main tasks: human attribution, profiling, human-vs-LLM detection, human-LLM coauthor detection, and LLM attribution. These tasks are applicable to both text and code, as shown in Figure~\ref{fig:enter-label}.

Human attribution involves identifying the individual author of a given piece of text or code. Profiling seeks to infer characteristics or traits of the author, such as demographics or writing style. Human-vs-LLM detection aims to determine whether content was produced by a human or an LLM. Human-LLM coauthor detection focuses on identifying instances where both a human and an LLM have collaborated on the same work. LLM attribution attempts to specify which language model generated the content. While AI-vs-human detection is well studied, attributing source code to specific LLMs remains unexplored. Analogous to tracing a photo back to its camera~\cite{dirik_source_2013,gupta_study_2021}, model-level attribution could enhance accountability, support academic integrity, and strengthen code security, especially since over 60\% of model-generated C code contains vulnerabilities~\cite{tihanyi_how_2024}. 

In this paper, we examine the attribution of C source code generated by various state-of-the-art LLMs, focusing our analysis on the following research questions:

\begin{tcolorbox}[
    colback=gray!10,    
    colframe=black,     
    arc=6pt,            
    boxrule=0.7pt,      
    left=2mm, right=2mm, top=1mm, bottom=1mm
    ]
\textbf{RQ1:} Can we perform authorship attribution on LLM-generated C code using arbitrary programming tasks?\\[0.5em]
\textbf{RQ2:} Which machine learning (ML) and transformer models perform the best for black-box C source code attribution?

\end{tcolorbox}

To answer these questions, we present four key contributions:

\begin{itemize}
  \item \textbf{CodeT5-Authorship:} We release a novel model based on the encoder layers of CodeT5, optimized for authorship classification. Our PyTorch implementation uses a two-layer classification head with GELU activation and dropout, enabling accurate attribution of C code to its source LLM.

  \item \textbf{LLM-AuthorBench:} We release a public dataset of, 32\,000 compilable C programs, each labeled by its source LLM, covering eight state-of-the-art models and diverse coding tasks. This benchmark serves as a standard reference for comparing various Transformer and machine learning models in the task of C code authorship attribution.

  \item \textbf{Comprehensive evaluation:} Using the newly released \textsc{LLM-AuthorBench} benchmark, we compare the \textsc{CodeT5-Authorship} model with traditional machine learning classifiers, like \textit{Random Forest (RF)}, \textit{XGBoost}, \textit{k-nearest neighbors (KNN)}, \textit{Support Vector Machine (SVM)}, and \textit{Decision Tree (DT)}, as well as eight fine-tuned transformer models: \textit{BERT, RoBERTa, CodeBERT, ModernBERT, DistilBERT, DeBERTa-V3, Longformer}, and a \textit{LoRA-optimized Qwen2-1.5B}.

  \item \textbf{High-accuracy attribution:} We demonstrate that model-level attribution is both feasible and accurate, achieving up to 97.56\% accuracy in binary classification within model families (e.g., GPT-4.1 vs.\ GPT-4o) and up to 95.40\% accuracy in multi-class attribution.

 \item \textbf{Open science:}  To support reproducibility, our dataset and the corresponding Google Colab training code are available on GitHub to facilitate future work in LLM authorship attribution: \url{https://github.com/LLMauthorbench}.
\end{itemize}

By shifting from simple AI-vs-human detection to precise model attribution for LLM-generated source code, we unlock new opportunities for accountability and security in software engineering. The rest of the paper is organized as follows. Section~\ref{sec:related_work} reviews related work. Section~\ref{sec:methodology} outlines the methodology used to construct the dataset and to train the classifiers. Section~\ref{sec:experimental_results} presents experimental results, Section~\ref{sec:limitations} discusses limitations and future work, while Section~\ref{sec:conclusion} concludes the paper.

\section{Related Work}
\label{sec:related_work}
First, this section examines techniques for authorship attribution, followed by an overview of the categories shown in Figure~\ref{fig:enter-label}.

\subsection{Attribution Techniques}
\subsubsection{Stylometry}
\label{sec:stylometry}
This method aims at the quantitative analysis of an author's low level stylistic traits to capture a unique ``coding fingerprint''. It relies on extracting features such as; (i) \textit{lexical attributes} (e.g., character and keyword frequencies, punctuation patterns, or common operands)~\cite{seroussi_authorship_2014}, (ii) \textit{layout and formatting habits} (e.g., use of whitespace, indentation style, comment patterns), or (iii) \textit{software‐engineering metrics} such as \textit{Cyclomatic Complexity} (CC)~\cite{McCabe}, depth-of-inheritance, or other static features~\cite{he_authorship_2024}. In~\cite{caliskan-islam_-anonymizing_2015}, Caliskan-Islam et al. presented the \textit{Code Stylometry Feature Set} (CSFS), developed specifically for code stylometry. Table~\ref{tab:feature-mapping} presents a comprehensive summary of different stylometric and other features that are frequently utilized in related work.

\subsubsection{Traditional Machine-Learning Methods}
Machine learning treats authorship as a supervised task: code is turned into high-dimensional feature vectors and a model predicts the author. Classic, inexpensive, and interpretable classifiers-logistic regression, naïve Bayes, SVMs, and simple ensembles-rely on hand-crafted features only \cite{he_authorship_2024}. Tree-based ensembles, such as Random Forest (RF) or XGBoost, add non-linear interactions and feature-importance rankings, yet still cannot learn new representations themselves, underscoring the need for automated representation learning.

\subsubsection{Graph-based representations}

This method captures an author's structural style by modeling code as graphs. Two key approaches include: (i) \textit{Abstract Syntax Trees} (ASTs), where features such as subtree shapes, node types, and AST-depth statistics are input to Graph Neural Networks to learn vector embeddings that capture syntactic style~\cite{guo_method_2022}; and (ii) \textit{Program Dependency Graphs} (PDGs) and \textit{Control-Flow Graphs} (CFGs), which encode execution flow and data dependencies---highlighting design choices like error handling or iteration patterns---and are resilient to obfuscation that affects only surface syntax~\cite{he_authorship_2024}.

\subsubsection{Dynamic and Behavioral Analysis}
Most methods rely on static code signals, yet compiling and running a program uncovers rich dynamic cues---memory allocation, resource usage, system calls, I/O patterns, performance metrics---that expose a developer's run-time habits and deeper algorithmic choices, and are harder to spoof with quick edits~\cite{wang_integration_2018}. Dynamic analysis, however, requires sandboxing untrusted or even non-compilable code, making it costly and impractical for routine use~\cite{9825799}. As a result, it remains largely confined to niches such as malware forensics~\cite{10.1145/3653973,7784595}.

\begin{table*}[t]
\scriptsize
\centering
\renewcommand{\arraystretch}{1.1}
\begin{tabularx}{\textwidth}{p{5cm}p{2.7cm}p{1.4cm}X}
\toprule
\textbf{Feature} & \textbf{Category} & \textbf{Text / Code} & \textbf{Typical Techniques} \\ \midrule
Byte / character $n$-gram frequencies~\cite{kim_marking_2025,frantzeskou_effective_2006} & Stylometric (Raw Lexical)      & Both & Language-agnostic slices of raw bytes/chars; robust to obfuscation; fed to SVM or random-forest baselines \\ 
\hline
Token $n$-gram frequencies (keywords, identifiers, operators \& symbols)~\cite{frantzeskou_source_2006} & Stylometric (Lexical) & Both & Contiguous token sequences ($n\!=\!1\text{–}5$) for SVM, naïve Bayes, CNN/RNN; unigram slice counts keywords and punctuation symbols \\ 
\hline
Average identifier length~\cite{caliskan-islam_-anonymizing_2015,caliskan_when_2018}                         & Stylometric (Lexical)          & Code & Mean characters per variable / function / class name; lightweight cue for RF/SVM, implicitly modelled by token-level RNN/CNN \\ 
\hline
Variable-naming conventions (casing, prefixes, suffixes)~\cite{quiring_misleading_2019}   & Stylometric (Lexical)        & Code & Ratios of camelCase : snake\_case, Hungarian prefixes (e.g., m\_), suffixes (\_t, \_ptr), digits/underscores; token features for RF/SVM or as side-inputs to transformers \\ 
\hline

AST node/bigram statistics (bag-of-nodes)~\cite{caliskan-islam_-anonymizing_2015,caliskan_when_2018} & Stylometric (Syntactic) & Code & Counts of AST node types, parent-child bigrams and shallow subtree sizes; robust to renaming/formatting; fed to RF/SVM baselines \\ \hline
AST path \& graph embeddings~\cite{alon_code2vec_2018,ullah_crolssim_2022} & Learned (Structural) & Code & Context-path multiset, PDG/CFG graphlets encoded by GGNN, GraphCodeBERT or UniXCoder; dense style vectors compared via cosine $k$-NN or light MLP \\ \hline

Code-complexity metrics (cyclomatic, LOC/statement-length, nesting depth)~\cite{tihanyi_how_2024,kim_marking_2025,thathsarani_comprehensive_2024} & Statistical (Structural) & Code & Classical software-engineering metrics delivered as numeric features to RF/SVM; complement AST cues \\ 
\hline
Whitespace, indentation \& brace style~\cite{quiring_misleading_2019} & Stylometric (Formatting)       & Code & Tab-vs-space ratio, brace placement (1TBS, Allman), trailing whitespace; strong hand-crafted signals in RF/SVM \\ 
\hline
Commenting style \& frequency~\cite{suresh_is_2025}                                   & Stylometric (Content)          & Code & Comment-code density, inline vs.\ block preference, docstring length; NLP on comments plus density counts in RF/SVM \\ 
\hline
API / library usage~\cite{kalgutkar_code_2019}                                            & Stylometric (Content)          & Both & One-hot or TF-IDF vectors of imported packages and fully qualified calls; deep variants use GNNs over call graphs \\ 
\hline
Control-flow patterns~\cite{wang_integration_2018}                                          & Stylometric (Sem./Struct.)     & Code & Loop-type distribution, recursion, switch vs.\ if-chains; extracted from CFG/AST and encoded for RF/GNN \\ 
\hline
Topic / semantic content~\cite{wang_integration_2018,alvarez-fidalgo_clave_2025}                                     & Stylometric (Semantic)         & Both & LDA topics on identifiers/comments or transformer embeddings; compared via cosine similarity or fed to MLP classifiers \\ 
\hline
Information-theoretic global metrics (entropy, LM perplexity, Zipf deviation, log-rank, intrinsic dimension)~\cite{choi_i_2025} & Statistical (Info-theoretic) & Both & Single-value corpus-level statistics plugged into ensemble detectors or used as anomaly thresholds \\ 
\hline
Lexical-richness metrics (TTR, Yule's $K$)~\cite{bayrami_code_2021, lisson_investigating_2018}                     & Stylometric + Statistical      & Both & Token-economy vs.\ repetition; Yule’s $K$ mitigates length bias of TTR; entered as scalar features in RF/SVM \\ 
\hline
Neural code-embedding vectors~\cite{alvarez-fidalgo_clave_2025, alsulami_source_2017, abuhamad_large-scale_2021} & Learned (Deep Representation)            & Code &Mean-pooled transformer embeddings (CodeBERT, GraphCodeBERT) or Siamese contrastive encoders; Dense vector signatures fine-tuned on style; compared with cosine $k$-NN or fed to lightweight MLPs \\ 
\hline
Opcode/idiom $n$-grams \& tool-chain fingerprints~\cite{caliskan_when_2018,abuhamad_large-scale_2021,song_binmlm_2022} & Stylometric (Binary) & Code & Opcode sequences, instruction idioms, call-graphlets and compiler-option artefacts from disassembly/decompilation; RF, RNN or mixture-of-experts classifiers \\ \hline
Runtime behavioural traces~\cite{wang_integration_2018} & Stylometric (Dynamic) & Code & System-call $n$-grams, memory-allocation and branch-coverage profiles captured in sandbox execution; fused with static features via RF/GNN for obfuscation resilience \\ 
\bottomrule
\end{tabularx}
\caption{Feature families for effective use in source-code authorship or origin attribution.}
\label{tab:feature-mapping}
\end{table*}

\subsubsection{Binary Code Authorship Attribution}
Attribution can target compiled binaries rather than source code~\cite{10.1145/3653973}, where Rosenblum et al.~\cite{rosenblum_who_2011} showed that stylistic features can survive compilation~\cite{he_authorship_2024}. 
Although accuracy is lower than with source code due to compiler-induced variability, de-anonymization remains feasible at rates far in excess than pure chance~\cite{he_authorship_2024}.

\subsubsection{Deep Learning and Neural Network Approaches}
Neural network models have advanced authorship attribution by automatically learning feature representations from code, eliminating manual feature engineering~\cite{abuhamad_code_2019}. Sequence-based architectures---like RNNs/LSTMs and CNNs, applied to token or character streams---capture hierarchical stylistic patterns and subtle cues, that manual feature engineering may miss~\cite{zafar_language_2020}. Although these models outperform classical methods as author counts grow, they require large labeled datasets to train effectively~\cite{zafar_language_2020}.

\subsubsection{Pre‑trained Language Models}
These models are trained on massive corpora of text using unsupervised or self-supervised methods, learning rich contextual embeddings that capture syntactic and semantic nuances far beyond traditional bag-of-words features. Generic text encoders such as BERT \cite{devlin-etal-2019-bert} and RoBERTa \cite{liu_roberta_2019} are now joined by code-aware variants---CodeBERT \cite{feng_codebert_2020}, GraphCodeBERT \cite{guo_graphcodebert_2021}, CodeT5 \cite{wang_codet5_2021}---that fuse token, AST, and data-flow cues. For authorship, their embeddings can be fine-tuned with a classifier or shaped via contrastive learning, using slanted triangular schedules and gradual unfreezing to curb catastrophic forgetting \cite{howard-ruder-2018-universal}. These models reach state-of-the-art accuracy with almost no manual features, but demand heavy compute, risk domain overfitting, are adversarially brittle, and still face tight context limits.

\subsubsection{Large Language Model-Based Attribution}
LLMs sidestep the heavy, label-hungry training of classical ML/DL pipelines by doing zero or few-shot attribution through in-context prompts \cite{brown_language_2020,choi_i_2025}. Because they learn universal code patterns, they transfer smoothly across languages and domains. In~\cite{choi_i_2025}, GPT-4 reaches 65-69\% accuracy on real-world datasets with 500 + authors using only one reference snippet per author, and shows some resistance to superficial obfuscation. Chain-of-thought prompting can even supply human-readable rationales \cite{wei_chain--thought_2022}. Practical hurdles remain-high compute cost, opaque decision logic, uncalibrated confidence, and steep usage fees-limiting LLMs in large-scale pipelines.

\subsection{Code Attribution Tasks and Benchmarks}

\subsubsection{Benchmarking Datasets}
Except for black-box detectors, most LLM-attribution methods still require carefully labeled corpora to surface stylistic cues. Distinguishing two models with highly different styles is trivial---for instance, a small 2-3 B-parameter model can be easily distinguished from a GPT-4-class model by their difference in C code complexity, length, or compilability. The task becomes difficult, however, as the number of models grows and stylistic footprints converge. Rigorous evaluation therefore demands a large, heterogeneous dataset that also covers closely related model families. 
Our goal is to test attribution among both closely related variants like GPT-4o and GPT-4.1, and different state-of-the-art model families. 
Existing datasets (Table \ref{tab:datasets}) fall short of our needs; therefore, we construct a new dataset featuring code generated by eight state-of-the-art LLMs to more effectively address our research questions.

\begin{table}[t]
\scriptsize
\centering
\setlength{\tabcolsep}{4pt}  
\renewcommand{\arraystretch}{1.05}
\begin{tabularx}{\columnwidth}{l p{2.1cm} X}
\toprule
\textbf{Dataset} & \textbf{Domain} & \textbf{Size / Notes}\\
\midrule
Google Code Jam & Contest code & 2008–2017; $\sim$12 k authors\\
Codeforces & Contest code & Hundreds of authors; multi-lang\\
GitHub OSS Corpus & Real-world repos & 500–1 k authors; diverse files\\
BigCloneBench & Java clones & 8 k methods; 6 k clone pairs\\
POJ-104 & Student OJ & 52 k C/C++ solutions\\
Karnalim Corpus & Plagiarism & 467 Java files (7 tasks)\\
APT Malware & APT binaries & 3.6 k samples; 12 groups\\
BCCC\_AuthAtt-24 & C++ authorship & 24 k files; 3 k authors\\
FormAIv2 & C vulnerabilities & 331,000 programs, 9 LLM authors \\
\bottomrule
\end{tabularx}
\caption{Potential datasets for code authorship. }
\label{tab:datasets}
\end{table}

\subsubsection{Human-to-Human Authorship Attribution}

Choi et al. showed that zero-shot prompting with LLMs can link two code fragments to the same author, but accuracy falls to around 65-69\% on large author sets \cite{choi_i_2025}. CLAVE, a deep model pretrained on 270 k GitHub Python files and fine-tuned on Code Jam data, boosts same-author detection to 90\% by learning a stylometric embedding space \cite{alvarez-fidalgo_clave_2025}. The current state of the art-an RNN plus random-forest ensemble-reaches 92\% accuracy across 8903 programmers \cite{abuhamad_large-scale_2021}. Despite these gains, the classifiers remain brittle: adversarial edits mislead both Abuhamad's and Caliskan-Islam's models in 99\% of attempts \cite{quiring_misleading_2019}. Table \ref{tab:humanresults} summarises the leading approaches.
\begin{table}[b]
\scriptsize
\centering
\renewcommand{\arraystretch}{1.1}
\begin{tabularx}{\columnwidth}{@{}p{2.9cm}p{1.9cm}p{0.9cm}p{0.9cm}@{}}
\toprule
\textbf{Paper (Year)} & \textbf{Dataset} & \textbf{\#\,Auth.} & \textbf{Acc.} \\
\midrule
Abuhamad et al.\ (2021)~\cite{abuhamad_large-scale_2021}     & GCJ + GitHub (C, C++, Java, and Python)          & 8,903  & 92.3\%\\

Abuhamad et al.\ (2018)~\cite{abuhamad_large-scale_2018}           & GCJ (all years)        & 1\,600 & 96\% \\
Caliskan--Islam et al.\ (2015)~\cite{caliskan-islam_-anonymizing_2015}  & GCJ (C/C++)            & 1\,600 & 92\% \\
Alsulami et al. (2017)~\cite{alsulami_source_2017}                   &   GCJ (python)            & 70    &  89\%     \\
Frantzeskou et al. (2006) ~\cite{frantzeskou_effective_2006}                 & 0SJava2                & 30  & 97\%\\

\bottomrule
\end{tabularx}
\caption{State-of-the-art human written code attribution}
\label{tab:humanresults}
\end{table}

\subsubsection{Author Profiling (Experience/Style Inference)}
This task mainly aims at assessing developers' skills, expertise and coding habits from their code~\cite{coskun_profiling_2022}. For instance, Bamidis et al.~\cite{bamidis_software_2024} fine-tuned CodeBERT to classify snippets by skill level and domain, while Dev2Vec derives embeddings from repository descriptions, issue histories, and API calls to quantify expertise~\cite{dakhel_dev2vec_2023}. In LLM profiling, most studies focus on code correctness~\cite{liu_is_2023} or security~\cite{khoury_how_2023,tihanyi_how_2024}, with code quality metrics largely overlooked. In~\cite{tihanyi_how_2024} Tihanyi et al. partially addressed this by measuring cyclomatic complexity over $331,000$ C samples from nine models, showing that lower cyclomatic complexity correlates with fewer vulnerabilities.

\subsubsection{AI or Human Classification}
Early work tried to repurpose text-based detectors such as \textsc{GPTZero}, OpenAI's classifier, and \textsc{GLTR}; however, several studies showed that these systems generalize poorly to source code
\cite{pan_assessing_2024,suh_empirical_2024}. Researchers therefore adapted ideas from text detection-perplexity analysis and zero-shot methods like \textsc{DetectGPT} \cite{mitchell_detectgpt_2023}. Xu and Sheng's \textsc{CodeVision} \cite{xu_codevision_2025} measures token-level language-model perplexity and detects ChatGPT-written homework solutions more reliably than naïve entropy checks. Nguyen et al.'s \textsc{GPTSniffer} \cite{nguyen_gptsniffer_2024} combines syntax-aware sampling with ensemble scoring, reaching F1~\(\approx\)~0.95
(\(96\,\%\) accuracy) for distinguishing human and ChatGPT code in C and
Python.  Across ten languages, Jin et al.\ \cite{10992332} show that a
RoBERTa-based binary classifier can still separate StarCoder2 code from
human code with \(84.1\,\%\pm3.8\,\%\) accuracy.  Choi et al.\
\cite{choi_i_2025} further demonstrates that zero- or few-shot GPT-4 can
attribute C++/Java code from 686 human authors with \(68.7\,\%\) top-1
accuracy, indicating strong built-in authorship cues.

\subsubsection{AI and Human Co-Authored Code Attribution}
Paek et al.~\cite{paek_detection_2025} study mixed repositories in Java and
report >\,96\% accuracy for spotting GPT-3.5/4 fragments, rising to 99\% when over 1 000 human authors are present. They pair each human snippet with four LLM-paraphrased versions, encode ten style cues, and show that their \textit{LPCodedec} classifier best separates ChatGPT rewrites and struggles most with WizardCoder.  \textit{CodeMirage}~\cite{guo_codemirage_2025} extends the one language setting to ten programming languages written by ten LLMs, achieving an F1 score of 0.95 in AI vs. human detection, but drops to 0.65 under cross-model paraphrasing, revealing the fragility of co-authored code detection.

\subsubsection{LLM-Generated Code Attribution}

Interest in tracing the authorship of LLM-generated code is rising, as shown by recent watermarking schemes for code \cite{li_resilient_2024,dathathri_scalable_2024}. Yet, systematic attribution studies are still rare. The closest effort, LPcode/LPcodedec \cite{park_detection_2025}, tackles multi-class attribution but only for \emph{LLM-paraphrased human code}, not for code produced outright by the models. 

\textbf{To our best knowledge, no prior work tackles large-scale attribution of \emph{genuinely LLM-generated} code across multiple model families}, a gap this research fills. Unlike human authors-whose programs often reveal distinctive stylistic cues (Table~\ref{tab:feature-mapping})--modern LLMs are trained on overlapping corpora drawn from millions of GitHub and Stack Overflow repositories. Because these models can emulate a vast spectrum of coding styles--including each other's-reliable attribution becomes significantly more ambiguous and technically demanding.

\begin{figure*}[t] 
\centering
\caption{The five-step methodology for LLM authorship attribution in C code.}
\includegraphics[width=1\textwidth]{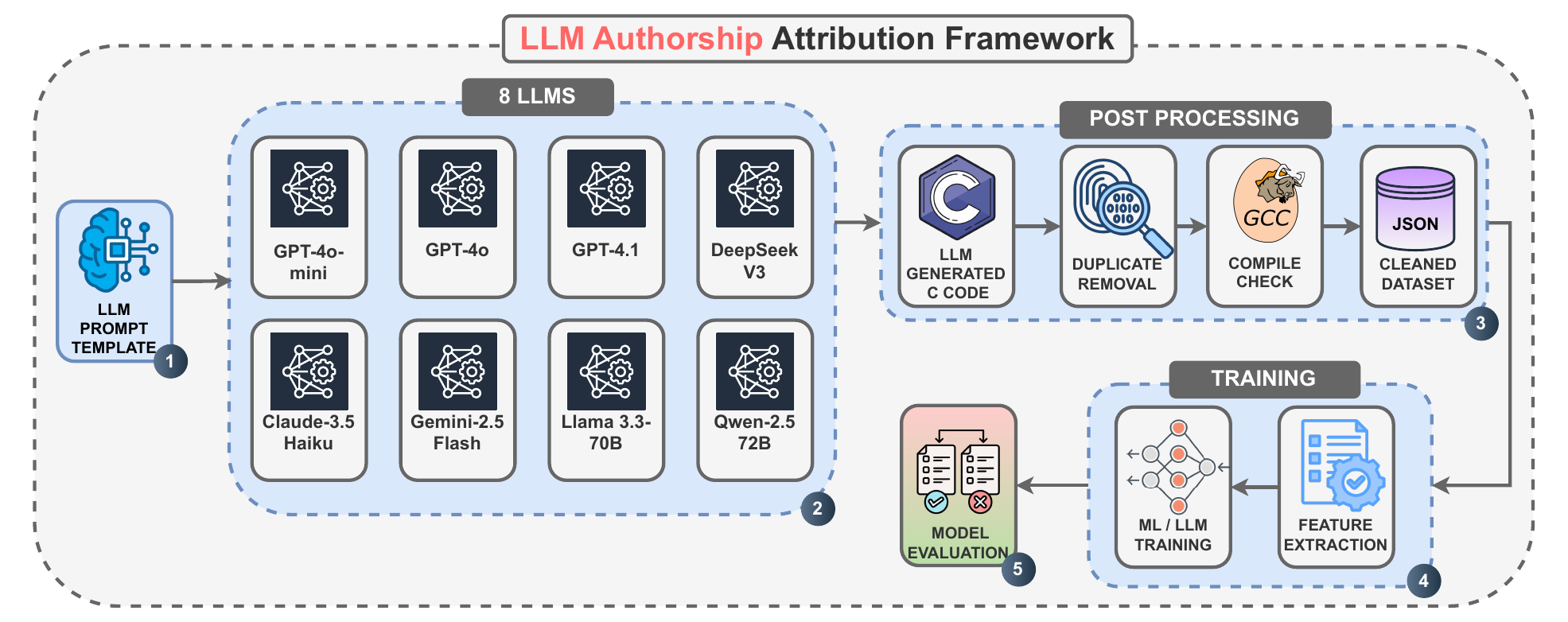} 
\label{fig:Framework}
\end{figure*}

\section{Methodology}

To evaluate LLM authorship attribution in C programming, we constructed a large, diverse benchmark, \textsc{LLM-AuthorBench} by 
\begin{enumerate}
    \item defining parameterized programming task templates;
    \item creating $4000$ different versions of the questions;
    \item prompting eight different LLMs to generate the C implementations of the tasks (32,000 in total);
    \item curating the final corpus through deduplication and training/validation splitting, where 80\% of the dataset is used for training, and 20\% is for validation.
\end{enumerate}
We evaluated the \textsc{CodeT5-Authorship} model alongside traditional machine learning and Transformer models to address our research questions. Figure~\ref{fig:Framework} outlines the main steps of our methodology, which are detailed in this section.

\label{sec:methodology}

\subsection{Dataset creation (LLM-AuthorBench)}

First, we manually created $N=300$ distinct templates, each describing a programming task with one or more variable parameters. Examples include:
\begin{itemize}
  \item \emph{Sorting:} ``Write a C program to sort an array of \texttt{\{size\}} integers using bubble sort.''
  \item \emph{Networking:} ``Create a C program that connects to server IP \texttt{\{ip\_address\}} on port \texttt{\{port\}} and sends the message \texttt{\{message\}}.''
\end{itemize}

Let $\mathcal{T} = \{ t_1, t_2, \ldots, t_N \}$ be the set of all question templates, where $N = |\mathcal{T}|$ is the total number of templates. For each template $t_i \in \mathcal{T}$, let $p_i$ denote the number of distinct questions that can be generated from $t_i$, i.e., $t_i \mapsto p_i$. 
Using the first example, $\text{size}$ is chosen randomly between $1$ and $100$. In this case, $t_1 \mapsto 100$, since there are $100$ possible distinct questions that can be generated from this template.

For templates that have multiple variable placeholders, $p_i$ is calculated as the product of the number of possible values for each placeholder in the template. For example, if $t_j$ has two placeholders, $\{a\}$ and $\{b\}$, with $n_a$ and $n_b$ possible values respectively, then $p_j = n_a \times n_b$. Thus, for the entire set,
\[
P = \sum_{i=1}^{N} \left( \prod_{k=1}^{K_i} n_{ik} \right)
\]
where $K_i$ is the number of placeholders in $t_i$ and $n_{ik}$ is the number of possible values for the $k$-th placeholder in $t_i$. For the $300$ distinct question templates, approximately $2.1$ billion unique questions can be generated. For this experiment, $4,000$ unique programming tasks were created which were given to $8$ state-of-the-art LLMs---GPT-4.1, GPT-4o, GPT-4o-mini, DeepSeek-v3, Qwen2.5-72B, Llama 3.3-70B, Claude-3.5-Haiku, and Gemini-2.5-Flash---yielding a total of $32,000$ C programs. We have ensured that only unique C programs are included in the final dataset.

\subsubsection{Deduplication}
Although each of the $4\,000$ questions is unique at the instantiation level, the underlying programming tasks (templates) repeat multiple times.  Concretely, each task template is provided to each LLM on average
$\frac{Q}{N} \;=\; \frac{4\,000}{300} \;\approx\; 13.3$ time. As a result, in \textsc{LLM-AuthorBench}, each programming task is implemented $13.3 \times 8 \;\approx\; 106.7$ times. Since LLMs rarely generate identical solutions, minor differences yield richer stylistic insights. We only remove fully identical (Type 0) programs and retain Type 1 and Type 2 clones~\cite{Cordy2011TheNC}, keeping variable renaming, formatting or comment changes, statement reordering-to capture each model's unique coding style. These nuanced distinctions have significantly contributed to our stylometric analysis.

\subsubsection{Compiler Validation}

We validated each sample using gcc with its \texttt{-c} flag enabled to allow for snippets that do not contain a main function. This removed invalid and incomplete code, as well as non-C code that was generated erroneously, to prevent them from polluting the training set. To balance the dataset after the deduplication and compile checks, we randomly dropped correct code snippets until we reached an equal number of programs for each model. At the time of writing, the total cost of generating the entire dataset was approximately \$350 USD (excluding model training, which utilized 2x A100 40 GB GPU), with GPT-4.1 and GPT-4o being the most expensive models.

\subsection{CodeT5-Authorship architecture}
In addition to benchmarking off-the-shelf traditional ML and generic Transformer models (Section \ref{sec:traditional}), we present a custom variant of CodeT5+, that is specifically tuned for code-authorship attribution. Because code attribution is a fine-grained classification problem, encoder-centric architectures are generally more beneficial, as we don't need the verbose output capabilities of decoder architectures. After exploring various design options, we selected the pretrained CodeT5+ 770M parameter sequence-to-sequence model and removed its decoder, resulting in a streamlined, encoder-only network optimized for attribution tasks. This modified CodeT5 variant consistently outperforms all tested models, including both encoder-only and decoder-only alternatives, in our experiments. Full results are presented in the next section.

\begin{figure}[ht] 
\centering
\includegraphics[width=0.30\textwidth]{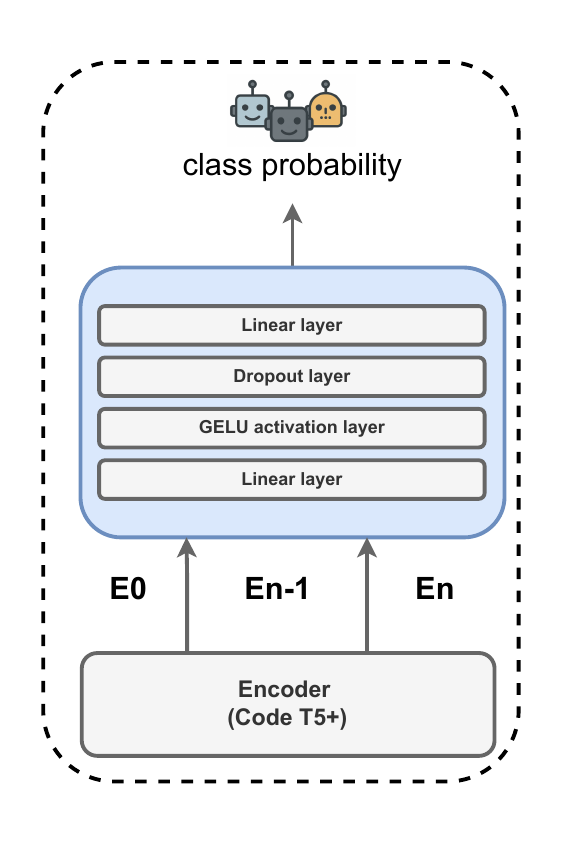} 
\caption{CodeT5-Authorship architecture}

\label{fig:CodeT5Authorship}
\end{figure}

As shown in Figure~\ref{fig:CodeT5Authorship} the encoder block is connected to a classification head that we implemented using the Pytorch library. The classification head is a sequence of two linear layers connected, with an activation layer and a dropout layer (20\%) in between. The first linear layer reduces the dimensionality of the last hidden layer output after taking the first token embedding. After experimenting with various activation functions, we selected GELU, which delivered the best performance for our model. The final linear layer then produces a probability vector assigning code attribution to each class.

\subsection{Traditional ML and Transformer Models}
\label{sec:traditional}
To identify the most effective approach for LLM authorship attribution without architectural modifications, first we evaluate both traditional machine learning baselines and modern Transformer-based architectures. Authorship attribution methods generally fall into two broad categories: (a) classical stylometric approaches that rely on engineered features and conventional classifiers, and (b) deep learning approaches leveraging neural language models~\cite{silva_forged-gan-bert_2024}. 

\begin{table*}[t]
\centering
\scriptsize
\caption{Selected file-level and function-level metrics used in our authorship-attribution experiments.}
\begin{tabularx}{\textwidth}{@{}L{0.23\textwidth}L{0.27\textwidth}X@{}}
\toprule
\textbf{Metric} & \textbf{Formula / Symbol} & \textbf{Description} \\
\midrule
\rowcolor{gray!20}
\multicolumn{3}{@{}l}{\textbf{Basic File Metrics}}\\
\hline
Character count  & $|\text{characters}|$                         & Total characters in the file, including whitespace and newline characters. \\
\hline
Line count       & $L$                                           & Total lines in the file (code, comments, and blank lines). \\
\hline
Function count   & $F_{\text{total}} = k$                        & Number of top-level (non-nested) functions defined in the file. \\
\hline
\rowcolor{gray!20}
\multicolumn{3}{@{}l}{\textbf{Function-Level Metrics (per function $f_j$)}}\\
\hline
Maximum nesting depth   & $D_{\max}(f_j)=\displaystyle\max_{n_i\in\text{AST}(f_j)}\text{depth}(n_i)$
                       & Deepest level of nested control structures within the function body.\\
\hline
AST node term frequency & $\text{tf}(t,f_j)=M(t,f_j)/N$ & Relative frequency of each AST node type $t$ in the function.\\
\hline
AST bigram frequency    & $\text{bigram\_tf}(t_a,t_b,f_j)= B(t_a\!\rightarrow\! t_b,f_j) /N$ 
                       & Relative frequency of ordered AST-node pairs $(t_a \!\rightarrow\! t_b)$ capturing local structure.\\
\hline
Cyclomatic complexity   & $CC(f_j)=\#\text{decision points}+1$   & Number of independent execution paths through the function’s control-flow graph.\\
\hline
Parameter count         & $P(f_j)=|\text{params}(f_j)|$          & Number of formal parameters the function accepts.\\
\hline
Lines of code (LOC)     & $\text{LOC}(f_j)=\bigl|\{s_i\mid s_i\in\text{exec.\ stmts of }f_j\}\bigr|$ 
                       & Executable statement lines in the function (excludes blank and comment lines).\\
\hline
Variable complexity     & $V(f_j)=|\{v_i\mid v_i\in\text{local vars of }f_j\}|$ 
                       & Count of distinct local variables declared in the function body.\\
\hline
Return count            & $R(f_j)=\sum \mathbb{I}_{\text{return}}$ 
                       & Number of explicit \texttt{return} statements in the function.\\
\hline
\rowcolor{gray!20}
\multicolumn{3}{@{}l}{\textbf{Halstead Metrics (per $f_j$)}}\\
\hline
Distinct operators      & $n_1$                                  & Unique operator tokens (e.g., \texttt{+}, \texttt{if}, \texttt{call}).\\
\hline
Distinct operands       & $n_2$                                  & Unique operand tokens (identifiers, literals).\\
\hline
Total operators         & $N_1$                                  & Total occurrences of operator tokens.\\
\hline
Total operands          & $N_2$                                  & Total occurrences of operand tokens.\\
\hline
Volume                  & $V=(N_1+N_2)\log_2(n_1+n_2)$           & Information content of the implementation (Halstead volume).\\
\hline
Difficulty              & $D=n_1/2 \cdot N_2/n_2$ & Estimated implementation difficulty.\\
\hline
Effort                  & $E=D\cdot V$                           & Estimated mental effort required to develop or comprehend the code.\\
\bottomrule
\end{tabularx}
\label{tab:features2}
\end{table*}

\subsubsection{ML Training and Feature Selection}
For classical machine-learning, we extracted all features listed in Table~\ref{tab:features2} using the Joern framework~\footnote{https://github.com/joernio/joern} and general-purpose natural language processing (NLP) algorithms. A central question is whether such manual feature engineering can rival, or even beat, approaches that use BERT-style Transformers as automatic feature extractors.
These metrics quantify structural, syntactic, and stylistic patterns inherent to programming styles. To clarify our notation for Table~\ref{tab:features2}, let $F$ denote a source-code file containing $k$ functions, $\{f_{1},\dots,f_{k}\}$.

\subsubsection{Transformer-based Approaches}

Many state-of-the-art models used in this comparison are based on the Transformer architecture, originally introduced by Vaswani et al.~\cite{vaswani_attention_2017}. The Transformer comprises encoder and decoder modules, and different models may utilize one or both depending on their design and task type. Encoder-based architectures, such as BERT, RoBERTa, CodeBERT, and Longformer, are primarily designed for representation learning and sequence understanding. \emph{Encoder-only} models like BERT employ bidirectional self attention, enabling every token to attend to both its left and right context and yielding rich, sentence-level representations that are well suited to classification tasks \cite{devlin-etal-2019-bert}. Their principal limitation is computational: the quadratic complexity of full self-attention restricts practical sequence length to roughly 512 sub-word tokens. \emph{Decoder-only} models (GPT style) rely on causal (left-to-right) attention and are pretrained for language generation. Recent variants---such as Qwen2---extend the usable context window to 32 thousand tokens or more, rendering them capable of processing entire source code files without truncation. 
To evaluate the performance of foundational models in our experiments, we fine-tune the following pretrained models:

\begin{itemize}[leftmargin=1.5em]

\item \textbf{BERT}~\cite{devlin-etal-2019-bert}.  
A 12-layer bidirectional encoder (110 M parameters) that looks at both left and right context, making it a solid baseline for sequence-level classification.  
Its primary limitation is the 512-token window imposed by quadratic self-attention.

\item \textbf{ModernBERT}~\cite{modernBERT}.
A major evolution of BERT, ModernBERT expands the original 512-token context window to up to 8,192 tokens. Alongside improved pre-training, positional encodings, and training methods, this enables stronger contextual representations and better performance-especially on code tasks—while keeping the encoder-only design.

\item \textbf{DistilBERT}~\cite{sanh_distilbert_2020}.  
A 6-layer, 66 M-parameter distillation of BERT that is 40\% smaller and 60\% faster at inference.  
It is ideal when GPU memory or latency is a bottleneck, but it inherits the same 512-token cap and lacks code-specific pre-training.

\item \textbf{RoBERTa}~\cite{liu_roberta_2019}.  
Keeps BERT’s architecture yet removes next-sentence prediction, applies dynamic masking, and is trained on a 10 times larger corpus.  
These changes systematically improve downstream accuracy, but the model is still encoder-only and restricted to 512 tokens.

\item \textbf{CodeBERT}~\cite{feng_codebert_2020}.  
Builds on the RoBERTa backbone and is jointly pre-trained on paired natural-language / source-code data (CodeSearchNet).  
This bimodal signal lets it capture lexical and structural properties of code, making it a promising candidate for authorship attribution compared to purely natural-language models, though the 512-token limit remains.

\item \textbf{Longformer}~\cite{beltagy_longformer_2020}.  
Introduces sliding-window self-attention with optional global tokens, reducing complexity from \(O(n^2)\) to \(O(n)\) and enabling context windows of up to $4096+$ tokens.  
It therefore processes entire C files that BERT-style models must truncate, while still benefiting from RoBERTa's pre-training.

\item \textbf{DeBERTa-v3}~\cite{he_debertav3_2023}.  
Adds disentangled content and position embeddings plus an ELECTRA-style replaced-token objective, yielding stronger representations than RoBERTa on many NLP tasks.  
Although it shares the 512-token ceiling, its finer positional modeling helps capture subtle stylistic cues in shorter code snippets.

\item \textbf{Qwen2-1.5B}~\cite{bai_qwen_2023}.  
A 1.5 billion parameter decoder-only LLM with a 32 k-token context window, pre-trained on a diverse corpus that includes code.  
Its unidirectional attention is less ``holistic'' than encoder models, but the massive context lets it ingest whole projects in one pass.  
We convert it into an 8-way classifier via \textbf{LoRA}, adding low-rank adapters \(\Delta W = AB\) (\(A\!\in\!\mathbb{R}^{d\times r},\,B\!\in\!\mathbb{R}^{r\times d},\,r\!\ll\!d\)), so only \(\approx\)1 \% of parameters are updated while the base weights stay frozen.

\end{itemize}

\section{Experimental Results}
\label{sec:experimental_results}

In this section, we explore the results of both binary classification and the more complex multi-class classification cases.

\subsection{Binary classification}

For the first experiment, our objective is to train a classifier on code acquired only from two LLMs, and subsequently achieve high accuracy for the attribution of C code unseen by the classifier model. To tackle the most challenging scenario, we used models belonging to the same architectural family, namely: \textit{GPT-4.1} and \textit{GPT-4o}. Given that these models are closely related, a natural question arises: \textit{Is it even possible to distinguish their generated code?}

\begin{figure}[b] 
\centering
\includegraphics[width=0.45\textwidth]{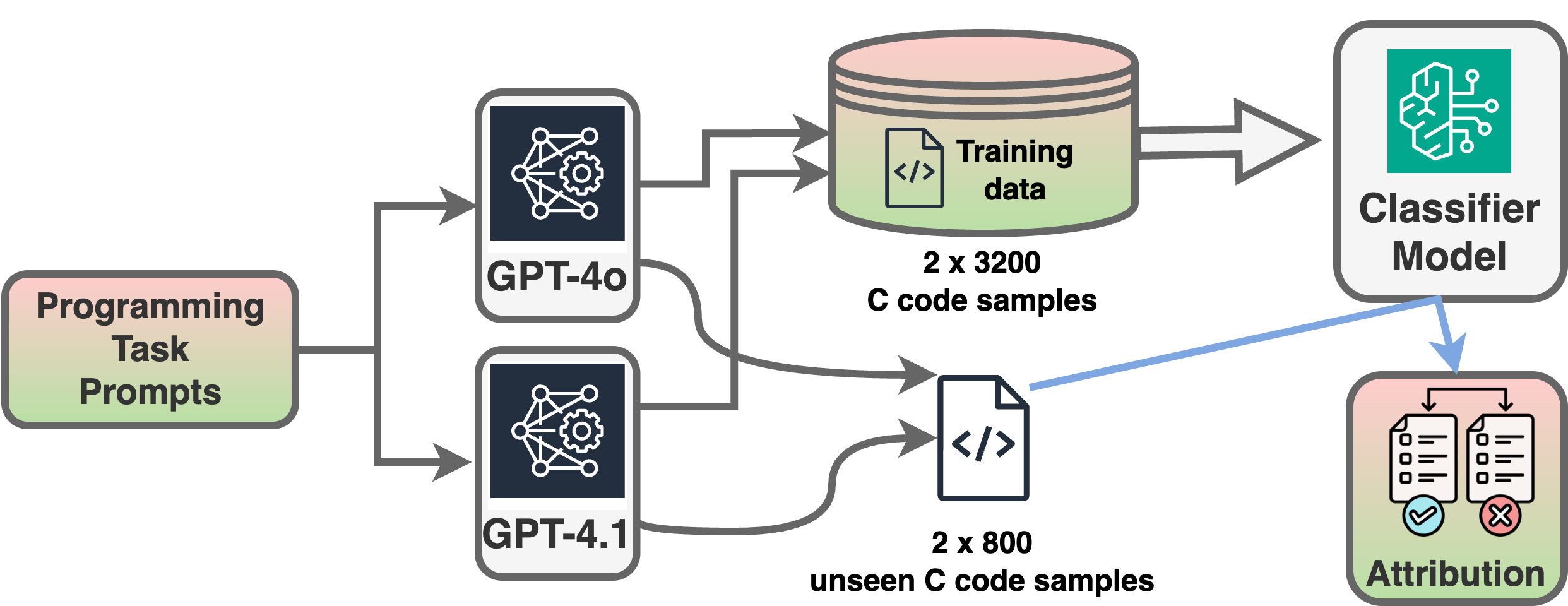} 
\caption{Binary classification: $50\% + \epsilon$?}

\label{fig:Framework2}
\end{figure}

Recent iterations of OpenAI's state-of-the-art non-reasoning models---\emph{GPT-4o} (April 2024) and the refreshed \emph{GPT-4.1} checkpoint (April 2025)--- were each pre-trained on a multi-trillion-token corpus that blends both text and code tokens, the latter being drawn from public GitHub repositories, Stack Overflow dumps, package registries etc.

While both models presumably share comparable pre-training corpora and compute budgets, we \emph{hypothesise} that their Supervised Fine-Tuning (SFT) and Reinforcement Learning from Human Feedback (RLHF) stages diverged: GPT-4o may have been aligned chiefly on mid-2023 data, whereas GPT-4.1 could incorporate a late-2024 GitHub snapshot and a richer pool of coding examples.
Such calibration differences might manifest in subtle stylistic signals, for example: comment density, brace positioning, whitespace regularity, and typical function granularity.  Empirically, we find these cues as evidenced by Listing~\ref{lst:gpt4_compare} where even a simple ``Hello World style'' program yielded different commenting styles, confirming that \emph{fine-tuning style drift} remains observable.

\begin{listing}[ht]
\centering

\begin{tcolorbox}[enhanced,
  title=GPT-4\textnormal{o} output,
  colframe=gray!50!black,colback=gray!10!white,
  arc=1mm,fonttitle=\bfseries,coltitle=black!50!black,
  attach boxed title to top text left={yshift=-0.50mm},
  boxed title style={
      skin=enhancedfirst jigsaw,size=small,arc=1mm,bottom=-1mm,
      interior style={fill=none,top color=gray!30!white,bottom color=gray!20!white}}]
\begin{minted}[fontsize=\scriptsize]{c}
#include <stdio.h>
int main() {
    // Print the message to the console
    printf("ACM AISec is an amazing conference!\n");
    return 0; // indicate successful execution
}
\end{minted}
\end{tcolorbox}

\vspace{-7pt} 

\begin{tcolorbox}[enhanced,
  title=GPT-4.1 output,
  colframe=gray!50!black,colback=gray!10!white,
  arc=1mm,fonttitle=\bfseries,coltitle=black!50!black,
  attach boxed title to top text left={yshift=-0.50mm},
  boxed title style={
      skin=enhancedfirst jigsaw,size=small,arc=1mm,bottom=-1mm,
      interior style={fill=none,top color=gray!30!white,bottom color=gray!20!white}}]
\begin{minted}[fontsize=\scriptsize]{c}
#include <stdio.h>
int main() {
    printf("ACM AISec is an amazing conference!\n");
    return 0;
}
\end{minted}
\end{tcolorbox}

\caption{Simple C programs showing the slightly more verbose commenting style in GPT-4\textnormal{o} compared to GPT-4.1.}
\label{lst:gpt4_compare}
\end{listing}

After extensive feature engineering experiments, we observed that \emph{none} of the structural metrics in Table~\ref{tab:features2} (e.g., AST or CFG counts) shifted the decision boundary of the classical models by more than $\approx$0.4\,pp.\ on the validation set.  In contrast, the presence or absence of \textit{comment tokens} moved most curves by 2-3\,pp., confirming that commenting style is a strong cue for LLM authorship. Table~\ref{tab:llm_ml_sorted} holds the results and yields four additional insights:

\begin{table}[t]
\centering
\scriptsize
\caption{Binary Classification: LLMs vs. Classical ML Models}
\renewcommand{\arraystretch}{1.4}
\sisetup{
    group-separator = {\,},
    group-minimum-digits = 4,
    output-decimal-marker = {.},
    table-number-alignment = right
}
\rowcolors{2}{gray!15}{white}
\begin{tabular}{l c S S c c l}
\toprule
\textbf{Model Name} & \textbf{Type} & \textbf{Acc (\%)} & \textbf{Prec (\%)} & \textbf{Time} & \textbf{Co.} & \textbf{Key Parameters} \\
\midrule
\rowcolor{green!20}CodeT5-Authorship      & LLM & 97.56 & 97.59 & 74:14 & \ding{52} & Layers: 24, Token: 512 \\
DeBERTa-V3      & LLM & 97.00 & 97.00 & 151:46 & \ding{52} & Layers: 12, Token: 2048 \\
QWEN2-1.5B      & LLM & 96.88 & 96.87 & 179:54 & \ding{52} & Layers: 32, Token: 2048 \\
DeBERTa-V3      & LLM & 96.75 & 96.81 & 45:21  & \ding{52} & Layers: 12, Token: 1024 \\
DeBERTa-V3      & LLM & 96.31 & 96.32 & 27:26  & \ding{52} & Layers: 12, Token: 512 \\
Longformer      & LLM & 96.19 & 96.19 & 117:42 & \ding{52} & Layers: 12, Token: 2048 \\
ModernBERT$_{B}$& LLM & 95.94 & 95.95 & 36:04  & \ding{52} & Layers: 12, Token: 512 \\
RoBERTa$_{L}$   & LLM & 95.68 & 95.76 & 87:35  & \ding{52} & Layers: 24, Token: 512 \\
codeBERT        & LLM & 95.31 & 95.43 & 30:21  & \ding{52} & Layers: 12, Token: 512 \\
RoBERTa$_{B}$   & LLM & 94.81 & 94.87 & 30:21  & \ding{52} & Layers: 12, Token: 512 \\
BERT$_{B}$      & LLM & 94.75 & 94.81 & 31:05  & \ding{52} & Layers: 12, Token: 512 \\
DistilBERT$_{B}$& LLM & 93.81 & 93.82 & 19:04  & \ding{52} & Layers: 6, Token: 512 \\
codeBERT        & LLM & 93.68 & 93.75 & 30:21  & \ding{56} & Layers: 12, Token: 512 \\
XGBoost         & ML  & 92.2  & 92.2  & 5.21   & \ding{52} & Estimators: 400 \\
RoBERTa$_{B}$   & LLM & 92.81 & 92.84 & 30:33  & \ding{56} & Layers: 12, Token: 512 \\
Random Forest   & ML  & 90.4  & 90.4  & 12.33  & \ding{52} & Estimators: 400 \\
BERT$_{B}$      & LLM & 91.62 & 91.69 & 30:24  & \ding{56} & Layers: 12, Token: 512 \\
DistilBERT$_{B}$& LLM & 91.00 & 91.09 & 18:19  & \ding{56} & Layers: 6, Token: 512 \\
XGBoost         & ML  & 89.7  & 89.7  & 5.98   & \ding{56} & Estimators: 400 \\
SVM (Kernel)    & ML  & 88.9  & 88.9  & 4.32   & \ding{52} & Kernel: RBF \\
Random Forest   & ML  & 88.2  & 88.3  & 11.49  & \ding{56} & Estimators: 400 \\
Bagging (DT)    & ML  & 84.9  & 84.9  & 6.41   & \ding{52} &  Estimators: 10 \\
Bagging (DT)    & ML  & 84.7  & 84.8  & 5.93   & \ding{56} & Estimators: 10 \\
SVM (Linear)    & ML  & 86.4  & 86.4  & 0.09   & \ding{52} & Max\_iter=2000 \\
SVM (Kernel)    & ML  & 84.2  & 84.3  & 4.98   & \ding{56} & Kernel: RBF \\
KNN             & ML  & 83.5  & 83.5  & 0.00   & \ding{52} & Neighbors: 5 \\
SVM (Linear)    & ML  & 80.6  & 80.6  & 0.10   & \ding{56} & Kernel: Linear \\
KNN             & ML  & 80.3  & 80.4  & 0.00   & \ding{56} & Neighbors: 5 \\
Decision Tree   & ML  & 77.1  & 77.1  & 0.39   & \ding{52} & Max Depth: 8 \\
Decision Tree   & ML  & 74.2  & 74.3  & 0.33   & \ding{56} & Max Depth: 8 \\
\bottomrule
\end{tabular}
Legend: $_{B}$: base model, Comm: Comment 
\label{tab:llm_ml_sorted}
\end{table}

\begin{enumerate}[label=(\roman*)]

    \item \textbf{CodeT5-Authorship wins the race:}  
          Our custom modified CodeT5 model at 97.56\% beats all off the shelf architectures. Most surprisingly, it does so, while only seeing truncated---has 512 token limit---code snippets, and it requires half as much training time as the second best model---DeBERTa-V3 with 97\%---which has 2048 token context window. 

    \item \textbf{LLMs dominate classical ML:}  
          Every LLM that keeps comments surpasses 93\,\%\,accuracy,
          whereas the best classical ML model
          (XGBoost with comments, 92.2\,\%) still trails CodeT5-Authorship by over \mbox{5\,pp.}

    \item \textbf{Comment removal hurts all models, but modestly:}  
          The median drop is $2.5$\,pp.\ for classical learners
          (e.g., XGBoost: 92.2\,$\rightarrow$\,89.7) and
          $3.1$\,pp.\ for LLMs (BERT$_B$: 94.75\,$\rightarrow$\,91.62),
          showing that stylistic signals in the source code itself remain exploitable.

    \item \textbf{In the binary task, long context yields diminishing returns:}  
      Expanding the window from 512 to 2\,048 tokens boosts DeBERTa-V3 by only +0.69 pp.\ (96.31 → 97.00 \%), and Longformer also trails CodeT5-Authorship by 1.37 pp. These results could imply that (i) the critical stylistic cues reside mostly in the first 512 tokens or (ii) architectural differences are valuable than sheer context length. 

    \item \textbf{Code-specific pre-training helps, but is not sufficient:}  
          CodeBERT with comments (95.31\%) marginally outperforms the general-purpose RoBERTa$_B$ (94.81\%). However, it falls short of models like DeBERTa-V3-512, ModernBERT$_{B}$, and even RoBERTa$_{L}$ which has the same context window, but has additinal layers, suggesting that while domain-specific pre-training is good, enhanced architecture is more important.
          
    \item \textbf{Small domain-tuned models can surpass larger,
          generic ones:}  
          Both the 2048 token DeBERTa-V3 (97.0\,\%), and CodeT5-Authorship outperforms the much larger
          QWEN2-1.5B, highlighting that parameter count alone is not a guarantee of top accuracy when architecture-level improvements are strong.
          
\end{enumerate}

\subsection{Cross-model multi-class attribution}
\label{sec:multiclass}

Table~\ref{tab:llm_ml_multiclass} shows accuracy and precision for multi-class attribution on C code generated by Gemini2.5 Flash, Claude3.5 Haiku, GPT-4.1, Llama 3.3, and DeepSeek-V3. For this part of the experiment, we decided to keep the comments, as we have already seen that it improves attribution efficacy. To gasp how significant it is in this setting, we tested the base BERT model with and without comments. We have gained the following insights:

\begin{table}[t]
\centering
\scriptsize
\caption{LLM vs Classical ML: Multi-class Classification}
\renewcommand{\arraystretch}{1.5}
\sisetup{
    group-separator = {\,},
    group-minimum-digits = 4,
    output-decimal-marker = {.},
    table-number-alignment = right
}
\rowcolors{2}{gray!15}{white}
\begin{tabular}{l c S S c c l}
\toprule
\textbf{Model Name} & \textbf{Type} & \textbf{Acc (\%)} & \textbf{Prec (\%)} & \textbf{Time} & \textbf{Co.} & \textbf{Key Parameters} \\
\midrule
\rowcolor{green!20} CodeT5-Authorship         & LLM & 95.40 & 95.41 & 185:55 & \ding{52} & Layers: 24, Token: 512 \\
Longformer         & LLM & 95.00 & 95.01 & 604:42 & \ding{52} & Layers: 12, Token: 2048 \\
DeBERTa-V3         & LLM & 94.25 & 94.32 & 107:02 & \ding{52} & Layers: 12, Token: 512 \\
DeBERTa-V3         & LLM & 94.15 & 94.28 & 733:32 & \ding{52} & Layers: 12, Token: 2048 \\
DeBERTa-V3         & LLM & 94.02 & 94.13 & 244:32 & \ding{52} & Layers: 12, Token: 1024 \\
codeBERT           & LLM & 93.52 & 93.64 & 80:40  & \ding{52} & Layers: 12, Token: 512 \\
RoBERTa$_{B}$      & LLM & 93.38 & 93.43 & 80:34  & \ding{52} & Layers: 12, Token: 512 \\
DistilBERT$_{B}$   & LLM & 93.02 & 93.06 & 54:58  & \ding{52} & Layers: 6, Token: 512 \\
BERT$_{B}$         & LLM & 92.65 & 92.71 & 85:05  & \ding{52} & Layers: 12, Token: 512 \\
QWEN2-1.5B         & LLM & 91.87 & 91.86 & 454:43 & \ding{52} & Layers: 32, Token: 2048 \\
XGBoost            & ML  & 90.80 & 90.80 & 00:57  & \ding{52} & Estimators: 400, Depth:9 \\
Random Forest      & ML  & 88.00 & 88.00 & 00:38  & \ding{52} & Estimators: 400 \\
BERT$_{B}$         & LLM & 85.45 & 85.79 & 80:38  & \ding{56} & Layers: 12, Token: 512 \\
SVM (Kernel)       & ML  & 81.40 & 81.40 & 00:40  & \ding{52} & Kernel: RBF \\
Bagging (DT)       & ML  & 78.40 & 78.70 & 00:19  & \ding{52} &  Estimators: 10 \\

SVM (Linear)       & ML  & 74.60 & 73.90 & 00:01  & \ding{52} & Max\_iter=2000\\
KNN                & ML  & 71.40 & 72.70 & 00:00  & \ding{52} & Neighbors: 5 \\
Decision Tree      & ML  & 58.90 & 61.20 & 00:01  & \ding{52} & Max Depth: 8 \\
\bottomrule
\end{tabular}
\textbf{Legend:} ${B}$: base model, Co: Comment present in C code
\label{tab:llm_ml_multiclass}
\end{table}

\begin{enumerate}[label=(\roman*)]
    \item \textbf{Long context still matters, but architecture matters more:}  
        Our CodeT5-Authorship model once again leads the results, followed closely by Longformer (12 layers, 2,048-token window), which achieves 95.0\% accuracy.

    \item \textbf{Surprising parameter efficiency of DeBERTa-V3:}  
          All three DeBERTa variants cluster tightly around
          94.0–94.3\%, and the \textit{shortest} window
          (512 tokens) actually edges out the 1\,024 and 2\,048-token
          versions, suggesting that once the model has gathered the cues it needs accuracy plateaus even if the window keeps growing. We note that while DeBERTa-v3 can process token sequences longer than 512, it is heavily optimized for a 512-token window.

    \item \textbf{Sheer LLM parameter size is not what matters:}  
          Several models, even including the simple base BERT model beats QWEN2-1.5B (1.5 B parameters), showing that capturing high-level style cues, and fine-tuning efficiency matter more than long-range generation ability.

    \item \textbf{Comments remain the dominant cue:}  
          Removing comments from BERT\(_B\) slashes accuracy from
          92.65\% to 85.45\% (a 7.2 pp.\ drop), nearly three times the
          average loss observed when ablating comments in the binary
          experiment. Hence, comment phrasing remains the single richest
          stylometric feature in the multi-class setting as well.

\item \textbf{Classical ML remains competitive for the price:}  
      A tuned XGBoost reaches \textbf{90.8\%} accuracy in \textbf{57 s},
      coming within 4.6 pp.\ of the best LLM (CodeT5-Authorship) while requiring \(\sim\!600\times\) less training time and no GPU acceleration.

\item \textbf{Relative gap between ML and LLMs persists as authors grow:}  
      The best ML model sits 4-5 pp.\ below Longformer in the five LLM author classification task, echoing the 5 pp.\ gap observed in the binary setting; LLMs therefore maintain a stable advantage as the number of candidate authors grows.

\end{enumerate}

\noindent

\begin{figure}[b]
    \centering
    \includegraphics[width=0.9\linewidth]{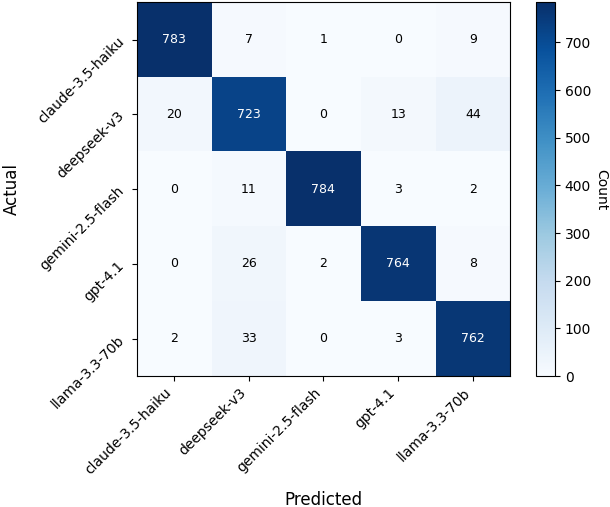}
    \caption{Confusion Matrix for CodeT5-Authorship}
    \label{fig:confusion}
\end{figure}
As Figure \ref{fig:confusion} shows, DeepSeek-V3 is far more often confused with other models than any of its peers. With CodeT5-Authorship, 723 of the 800 DeepSeek cases are classified correctly; most of the rest are mis-labelled as Llama-3.3-70B, Claude-3.5-haiku, or GPT-4.1. Similarly, true GPT-4.1 and Llama-3.3-70B outputs are frequently tagged as DeepSeek.

\begin{figure}[b]
    \centering
    \includegraphics[width=0.9\linewidth]{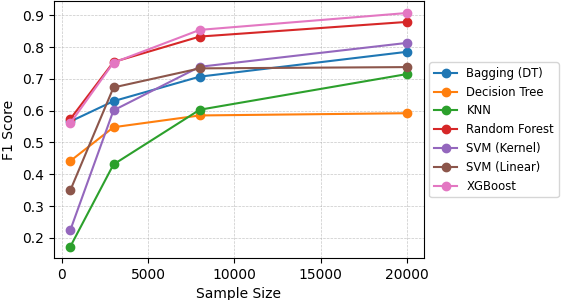}
    \caption{Classifier F1 score vs. sample size}
    \label{fig:f1}
\end{figure}

Figure~\ref{fig:f1} shows that for the five way classification, instead of using 4000 C programs per mode, approximately 2000 would have been sufficient to archive similar levels of accuracy.

In summary, across different model families, author attribution is highly feasible. These findings show that stylistic fingerprints persist despite vendor-specific pre-training pipelines, and that they remain detectable by off-the-shelf language encoders.

\subsection{Cross-check validation}

All samples in the \textsc{LLM-Authorship} benchmark were first generated through the \textit{OpenRouter.ai}\footnote{\url{https://openrouter.ai}} API (GPT-4.1, GPT-4o, and GPT-4o-mini). To verify that our classifier does not depend on proxy-specific artifacts, we built an out-of-distribution test set comprising code obtained directly from the native OpenAI API and produced from entirely new prompts and templates absent from the training corpus. One such prompt yields the illustrative C programs in Listing~\ref{lst:gpt4_compare}. On this unseen material, the classifier still attributed GPT-4o and GPT-4.1 with 99\,\% and 100\,\% confidence, respectively, confirming that it captures intrinsic stylistic cues of each LLM rather than OpenRouter watermarks or memorized examples.

\section{Limitations and Future Research}

\label{sec:limitations}
We have identified the following limitations, along with open research questions, that would be an interesting avenue for future research:

\begin{itemize}
    \item We focused on moderate-sized Transformers and did not train models larger than 7B parameters; scaling to that regime would entail distributed training across many high-memory GPUs, specialized optimization frameworks, and compute budgets that exceed the scope of this study.
    \item In contrast to previous human-attribution studies, which often involve many authors, a promising direction for future work would be to include additional LLMs in the multi-class attribution experiment to examine how performance changes as the number of models increases;
    \item The current training corpus is limited to C source code. It is therefore unclear whether the same stylistic signals transfer to other languages such as C++, Rust, Python or Java, or whether cross-language attribution (e.g., training on C and testing on C++) is feasible. Building multilingual author benchmarks and evaluating both intra and cross-language performance would fill this gap;
    \item We did not test robustness against deliberate style obfuscation. Stress-testing the classifiers under such noise and adversarial paraphrasing is crucial for assessing real-world deployability;
    \item Our experiment only focuses on a closed-group attribution, but an evaluation in which code generated by unseen LLMs (from the classifier's perceptive) are treated as out-of-distribution authors would be an interesting next step. Future work should also examine attribution accuracy on reasoning-centric LLMs and study whether they produce code that can strengthen or dilute the authorship signal;
    \item We did not evaluate large decoder-only LLMs as zero or few-shot classifiers as this approach requires prompt design rather than fine-tuning; therefore we leave this topic as an open research question for future work.

    \item Finally, the most effective long-context models require a lot of GPU-time, which may be prohibitive once the author set becomes massive. Future research should consider distillation, sparse adapters or other efficiency techniques to reduce the computational and energy cost of large-scale attribution.

\end{itemize}

\section{Conclusion}
\label{sec:conclusion}

In this paper, we released \textsc{LLM-AuthorBench}, a corpus of, 32\,000 compilable C programs produced by eight large recent language models over a broad selection of programming tasks. On this benchmark, we compared (i) seven classical ML classifiers that exploit lexical, syntactic, and structural features and (ii) eight fine-tuned Transformers: BERT, RoBERTa, CodeBERT, ModernBERT, DistilBERT, DeBERTa-V3, Longformer and LoRA-fine-tuned Qwen2-1.5B.

DeepSeek-V3 was misclassified more often than any other model. CodeT5-Authorship correctly identifies only 723 of 800 DeepSeek samples; the remainder are labelled as Llama-3.3-70B, Claude-3.5-haiku or GPT-4.1. The reverse is also true: GPT-4.1 and Llama-3.3-70B outputs are frequently tagged as DeepSeek.

\begin{tcolorbox}[colback=gray!10]
\begin{itemize}
    \item \textbf{RQ1:} Can we perform authorship attribution on \emph{LLM-generated} C code for arbitrary programming tasks? 
    \item \textbf{Answer: Yes, both in binary classification and in a five model multi-class scenario.} \\ Our custom model, \textsc{CodeT5-Authorship} encoder (24 layers, 512-token window) achieves \textbf{97.6\,\% accuracy on binary classification}, surpassing all baseline models. In the harder \textbf{multi-class scenario} it again edges out (\textbf{96 \%}), the best long-context baseline (Longformer, 95 \%).  Removing comments lowers accuracy by only 2-3 pp., confirming that stylistic cues are important for reliable attribution.
\end{itemize}
\end{tcolorbox}

\vspace{-1em}

\begin{tcolorbox}[colback=gray!10]
\begin{itemize}
    \item \textbf{RQ2:} Which machine-learning and Transformer models perform best for black-box C-source attribution?
    \item \textbf{Answer:} \textbf{Transformer encoders dominate.}  \\
      Both in the binary experiment and the multi-class setting, Transformer encoders outperform traditional ML. \textsc{CodeT5-Authorship} (24-layer encoder, 512 tokens) attains the \textbf{top accuracy of 97.6\%}, outpacing the best generic encoder DeBERTa-V3 by 0.6 pp.\ while using half the GPU time.  The strongest classical model, XGBoost, lags by more than 5 pp.  Decoder-only LLMs (e.g., Qwen2-1.5B) and longer context windows offer only marginal gains, indicating that architecture is more important than sheer parameter count or context length.
    \end{itemize}
\end{tcolorbox}

In summary, our results demonstrate that reliable authorship attribution of LLM-generated C code is feasible with moderately sized Transformer encoders, and traditional machine learning classifier alike. 
Looking ahead, there are several interesting research questions that can be investigated as highlighted in Section~\ref{sec:limitations} such as extending attribution to larger author sets, additional programming languages, zero/few-shot scenarios with foundation models, and adversarial obfuscation robustness will be critical for understanding the ultimate limits and practical utility of automated source-code authorship analysis.

\bibliographystyle{ACM-Reference-Format}
\bibliography{main}


\begin{thebibliography}{67}


\ifx \showCODEN    \undefined \def \showCODEN     #1{\unskip}     \fi
\ifx \showISBNx    \undefined \def \showISBNx     #1{\unskip}     \fi
\ifx \showISBNxiii \undefined \def \showISBNxiii  #1{\unskip}     \fi
\ifx \showISSN     \undefined \def \showISSN      #1{\unskip}     \fi
\ifx \showLCCN     \undefined \def \showLCCN      #1{\unskip}     \fi
\ifx \shownote     \undefined \def \shownote      #1{#1}          \fi
\ifx \showarticletitle \undefined \def \showarticletitle #1{#1}   \fi
\ifx \showURL      \undefined \def \showURL       {\relax}        \fi
\providecommand\bibfield[2]{#2}
\providecommand\bibinfo[2]{#2}
\providecommand\natexlab[1]{#1}
\providecommand\showeprint[2][]{arXiv:#2}

\bibitem[Abuhamad et~al\mbox{.}(2018)]%
        {abuhamad_large-scale_2018}
\bibfield{author}{\bibinfo{person}{Mohammed Abuhamad}, \bibinfo{person}{Tamer AbuHmed}, \bibinfo{person}{Aziz Mohaisen}, {and} \bibinfo{person}{DaeHun Nyang}.} \bibinfo{year}{2018}\natexlab{}.
\newblock \showarticletitle{Large-{Scale} and {Language}-{Oblivious} {Code} {Authorship} {Identification}}. In \bibinfo{booktitle}{\emph{Proceedings of the 2018 {ACM} {SIGSAC} {Conference} on {Computer} and {Communications} {Security}}} \emph{(\bibinfo{series}{{CCS} '18})}. \bibinfo{publisher}{Association for Computing Machinery}, \bibinfo{address}{New York, NY, USA}, \bibinfo{pages}{101--114}.
\newblock
\showISBNx{978-1-4503-5693-0}
\href{https://doi.org/10.1145/3243734.3243738}{doi:\nolinkurl{10.1145/3243734.3243738}}


\bibitem[Abuhamad et~al\mbox{.}(2021)]%
        {abuhamad_large-scale_2021}
\bibfield{author}{\bibinfo{person}{Mohammed Abuhamad}, \bibinfo{person}{Tamer Abuhmed}, \bibinfo{person}{David Mohaisen}, {and} \bibinfo{person}{Daehun Nyang}.} \bibinfo{year}{2021}\natexlab{}.
\newblock \showarticletitle{Large-scale and {Robust} {Code} {Authorship} {Identification} with {Deep} {Feature} {Learning}}.
\newblock \bibinfo{journal}{\emph{ACM Trans. Priv. Secur.}} \bibinfo{volume}{24}, \bibinfo{number}{4} (\bibinfo{date}{July} \bibinfo{year}{2021}), \bibinfo{pages}{23:1--23:35}.
\newblock
\showISSN{2471-2566}
\href{https://doi.org/10.1145/3461666}{doi:\nolinkurl{10.1145/3461666}}


\bibitem[Abuhamad et~al\mbox{.}(2019)]%
        {abuhamad_code_2019}
\bibfield{author}{\bibinfo{person}{Mohammed Abuhamad}, \bibinfo{person}{Ji-su Rhim}, \bibinfo{person}{Tamer AbuHmed}, \bibinfo{person}{Sana Ullah}, \bibinfo{person}{Sanggil Kang}, {and} \bibinfo{person}{DaeHun Nyang}.} \bibinfo{year}{2019}\natexlab{}.
\newblock \showarticletitle{Code authorship identification using convolutional neural networks}.
\newblock \bibinfo{journal}{\emph{Future Generation Computer Systems}}  \bibinfo{volume}{95} (\bibinfo{date}{June} \bibinfo{year}{2019}), \bibinfo{pages}{104--115}.
\newblock
\showISSN{0167-739X}
\href{https://doi.org/10.1016/j.future.2018.12.038}{doi:\nolinkurl{10.1016/j.future.2018.12.038}}


\bibitem[Alon et~al\mbox{.}(2018)]%
        {alon_code2vec_2018}
\bibfield{author}{\bibinfo{person}{Uri Alon}, \bibinfo{person}{Meital Zilberstein}, \bibinfo{person}{Omer Levy}, {and} \bibinfo{person}{Eran Yahav}.} \bibinfo{year}{2018}\natexlab{}.
\newblock \bibinfo{title}{code2vec: {Learning} {Distributed} {Representations} of {Code}}.
\newblock
\href{https://doi.org/10.48550/arXiv.1803.09473}{doi:\nolinkurl{10.48550/arXiv.1803.09473}}
\newblock
\shownote{arXiv:1803.09473 [cs]}.


\bibitem[Alsulami et~al\mbox{.}(2017)]%
        {alsulami_source_2017}
\bibfield{author}{\bibinfo{person}{Bander Alsulami}, \bibinfo{person}{Edwin Dauber}, \bibinfo{person}{Richard Harang}, \bibinfo{person}{Spiros Mancoridis}, {and} \bibinfo{person}{Rachel Greenstadt}.} \bibinfo{year}{2017}\natexlab{}.
\newblock \showarticletitle{Source {Code} {Authorship} {Attribution} {Using} {Long} {Short}-{Term} {Memory} {Based} {Networks}}. In \bibinfo{booktitle}{\emph{Computer {Security} – {ESORICS} 2017}}, \bibfield{editor}{\bibinfo{person}{Simon~N. Foley}, \bibinfo{person}{Dieter Gollmann}, {and} \bibinfo{person}{Einar Snekkenes}} (Eds.). \bibinfo{publisher}{Springer International Publishing}, \bibinfo{address}{Cham}, \bibinfo{pages}{65--82}.
\newblock
\showISBNx{978-3-319-66402-6}
\href{https://doi.org/10.1007/978-3-319-66402-6_6}{doi:\nolinkurl{10.1007/978-3-319-66402-6_6}}


\bibitem[Bai et~al\mbox{.}(2023)]%
        {bai_qwen_2023}
\bibfield{author}{\bibinfo{person}{Jinze Bai}, \bibinfo{person}{Shuai Bai}, \bibinfo{person}{Yunfei Chu}, \bibinfo{person}{Zeyu Cui}, \bibinfo{person}{Kai Dang}, \bibinfo{person}{Xiaodong Deng}, \bibinfo{person}{Yang Fan}, \bibinfo{person}{Wenbin Ge}, \bibinfo{person}{Yu Han}, \bibinfo{person}{Fei Huang}, \bibinfo{person}{Binyuan Hui}, \bibinfo{person}{Luo Ji}, \bibinfo{person}{Mei Li}, \bibinfo{person}{Junyang Lin}, \bibinfo{person}{Runji Lin}, \bibinfo{person}{Dayiheng Liu}, \bibinfo{person}{Gao Liu}, \bibinfo{person}{Chengqiang Lu}, \bibinfo{person}{Keming Lu}, \bibinfo{person}{Jianxin Ma}, \bibinfo{person}{Rui Men}, \bibinfo{person}{Xingzhang Ren}, \bibinfo{person}{Xuancheng Ren}, \bibinfo{person}{Chuanqi Tan}, \bibinfo{person}{Sinan Tan}, \bibinfo{person}{Jianhong Tu}, \bibinfo{person}{Peng Wang}, \bibinfo{person}{Shijie Wang}, \bibinfo{person}{Wei Wang}, \bibinfo{person}{Shengguang Wu}, \bibinfo{person}{Benfeng Xu}, \bibinfo{person}{Jin Xu}, \bibinfo{person}{An Yang}, \bibinfo{person}{Hao Yang},
  \bibinfo{person}{Jian Yang}, \bibinfo{person}{Shusheng Yang}, \bibinfo{person}{Yang Yao}, \bibinfo{person}{Bowen Yu}, \bibinfo{person}{Hongyi Yuan}, \bibinfo{person}{Zheng Yuan}, \bibinfo{person}{Jianwei Zhang}, \bibinfo{person}{Xingxuan Zhang}, \bibinfo{person}{Yichang Zhang}, \bibinfo{person}{Zhenru Zhang}, \bibinfo{person}{Chang Zhou}, \bibinfo{person}{Jingren Zhou}, \bibinfo{person}{Xiaohuan Zhou}, {and} \bibinfo{person}{Tianhang Zhu}.} \bibinfo{year}{2023}\natexlab{}.
\newblock \bibinfo{title}{Qwen {Technical} {Report}}.
\newblock
\href{https://doi.org/10.48550/arXiv.2309.16609}{doi:\nolinkurl{10.48550/arXiv.2309.16609}}
\newblock
\shownote{arXiv:2309.16609 [cs]}.


\bibitem[Bamidis et~al\mbox{.}(2024)]%
        {bamidis_software_2024}
\bibfield{author}{\bibinfo{person}{Dimitris Bamidis}, \bibinfo{person}{Ilias Kalouptsoglou}, \bibinfo{person}{Apostolos Ampatzoglou}, {and} \bibinfo{person}{Alexandros Chatzigeorgiou}.} \bibinfo{year}{2024}\natexlab{}.
\newblock \showarticletitle{Software {Skills} {Identification}: {A} {Multi}-{Class} {Classification} on {Source} {Code} {Using} {Machine} {Learning}}.
\newblock \bibinfo{journal}{\emph{Global Clinical Engineering Journal}} \bibinfo{volume}{6}, \bibinfo{number}{SI6} (\bibinfo{date}{Dec.} \bibinfo{year}{2024}), \bibinfo{pages}{74--77}.
\newblock
\showISSN{2578-2762}
\href{https://doi.org/10.31354/globalce.v6iSI6.278}{doi:\nolinkurl{10.31354/globalce.v6iSI6.278}}


\bibitem[Bayrami and Rice(2021)]%
        {bayrami_code_2021}
\bibfield{author}{\bibinfo{person}{Parinaz Bayrami} {and} \bibinfo{person}{Jacqueline~E. Rice}.} \bibinfo{year}{2021}\natexlab{}.
\newblock \showarticletitle{Code {Authorship} {Attribution} using content-based and non-content-based features}. In \bibinfo{booktitle}{\emph{2021 {IEEE} {Canadian} {Conference} on {Electrical} and {Computer} {Engineering} ({CCECE})}}. \bibinfo{pages}{1--6}.
\newblock
\href{https://doi.org/10.1109/CCECE53047.2021.9569061}{doi:\nolinkurl{10.1109/CCECE53047.2021.9569061}}
\newblock
\shownote{ISSN: 2576-7046}.


\bibitem[Beltagy et~al\mbox{.}(2020)]%
        {beltagy_longformer_2020}
\bibfield{author}{\bibinfo{person}{Iz Beltagy}, \bibinfo{person}{Matthew~E. Peters}, {and} \bibinfo{person}{Arman Cohan}.} \bibinfo{year}{2020}\natexlab{}.
\newblock \bibinfo{title}{Longformer: {The} {Long}-{Document} {Transformer}}.
\newblock
\href{https://doi.org/10.48550/arXiv.2004.05150}{doi:\nolinkurl{10.48550/arXiv.2004.05150}}
\newblock
\shownote{arXiv:2004.05150 [cs]}.


\bibitem[Boenninghoff et~al\mbox{.}(2019)]%
        {boenninghoff_explainable_2019}
\bibfield{author}{\bibinfo{person}{Benedikt Boenninghoff}, \bibinfo{person}{Steffen Hessler}, \bibinfo{person}{Dorothea Kolossa}, {and} \bibinfo{person}{Robert~M. Nickel}.} \bibinfo{year}{2019}\natexlab{}.
\newblock \showarticletitle{Explainable {Authorship} {Verification} in {Social} {Media} via {Attention}-based {Similarity} {Learning}}. \bibinfo{publisher}{IEEE Computer Society}, \bibinfo{pages}{36--45}.
\newblock
\showISBNx{978-1-7281-0858-2}
\href{https://doi.org/10.1109/BigData47090.2019.9005650}{doi:\nolinkurl{10.1109/BigData47090.2019.9005650}}


\bibitem[Brown et~al\mbox{.}(2020)]%
        {brown_language_2020}
\bibfield{author}{\bibinfo{person}{Tom Brown}, \bibinfo{person}{Benjamin Mann}, \bibinfo{person}{Nick Ryder}, \bibinfo{person}{Melanie Subbiah}, \bibinfo{person}{Jared~D Kaplan}, \bibinfo{person}{Prafulla Dhariwal}, \bibinfo{person}{Arvind Neelakantan}, \bibinfo{person}{Pranav Shyam}, \bibinfo{person}{Girish Sastry}, \bibinfo{person}{Amanda Askell}, \bibinfo{person}{Sandhini Agarwal}, \bibinfo{person}{Ariel Herbert-Voss}, \bibinfo{person}{Gretchen Krueger}, \bibinfo{person}{Tom Henighan}, \bibinfo{person}{Rewon Child}, \bibinfo{person}{Aditya Ramesh}, \bibinfo{person}{Daniel Ziegler}, \bibinfo{person}{Jeffrey Wu}, \bibinfo{person}{Clemens Winter}, \bibinfo{person}{Chris Hesse}, \bibinfo{person}{Mark Chen}, \bibinfo{person}{Eric Sigler}, \bibinfo{person}{Mateusz Litwin}, \bibinfo{person}{Scott Gray}, \bibinfo{person}{Benjamin Chess}, \bibinfo{person}{Jack Clark}, \bibinfo{person}{Christopher Berner}, \bibinfo{person}{Sam McCandlish}, \bibinfo{person}{Alec Radford}, \bibinfo{person}{Ilya Sutskever}, {and}
  \bibinfo{person}{Dario Amodei}.} \bibinfo{year}{2020}\natexlab{}.
\newblock \showarticletitle{Language {Models} are {Few}-{Shot} {Learners}}. In \bibinfo{booktitle}{\emph{Advances in {Neural} {Information} {Processing} {Systems}}}, Vol.~\bibinfo{volume}{33}. \bibinfo{publisher}{Curran Associates, Inc.}, \bibinfo{pages}{1877--1901}.
\newblock
\urldef\tempurl%
\url{https://papers.nips.cc/paper/2020/hash/1457c0d6bfcb4967418bfb8ac142f64a-Abstract.html}
\showURL{%
\tempurl}


\bibitem[Caliskan et~al\mbox{.}(2018)]%
        {caliskan_when_2018}
\bibfield{author}{\bibinfo{person}{Aylin Caliskan}, \bibinfo{person}{Fabian Yamaguchi}, \bibinfo{person}{Edwin Dauber}, \bibinfo{person}{Richard Harang}, \bibinfo{person}{Konrad Rieck}, \bibinfo{person}{Rachel Greenstadt}, {and} \bibinfo{person}{Arvind Narayanan}.} \bibinfo{year}{2018}\natexlab{}.
\newblock \showarticletitle{When {Coding} {Style} {Survives} {Compilation}: {De}-anonymizing {Programmers} from {Executable} {Binaries}}. In \bibinfo{booktitle}{\emph{Proceedings 2018 {Network} and {Distributed} {System} {Security} {Symposium}}}. \bibinfo{publisher}{Internet Society}, \bibinfo{address}{San Diego, CA}.
\newblock
\showISBNx{978-1-891562-49-5}
\href{https://doi.org/10.14722/ndss.2018.23304}{doi:\nolinkurl{10.14722/ndss.2018.23304}}


\bibitem[Caliskan-Islam et~al\mbox{.}(2015)]%
        {caliskan-islam_-anonymizing_2015}
\bibfield{author}{\bibinfo{person}{Aylin Caliskan-Islam}, \bibinfo{person}{Richard Harang}, \bibinfo{person}{Andrew Liu}, \bibinfo{person}{Arvind Narayanan}, \bibinfo{person}{Clare Voss}, \bibinfo{person}{Fabian Yamaguchi}, {and} \bibinfo{person}{Rachel Greenstadt}.} \bibinfo{year}{2015}\natexlab{}.
\newblock \showarticletitle{De-anonymizing {Programmers} via {Code} {Stylometry}}. \bibinfo{pages}{255--270}.
\newblock
\showISBNx{978-1-939133-11-3}
\urldef\tempurl%
\url{https://www.usenix.org/conference/usenixsecurity15/technical-sessions/presentation/caliskan-islam}
\showURL{%
\tempurl}


\bibitem[Choi et~al\mbox{.}(2025)]%
        {choi_i_2025}
\bibfield{author}{\bibinfo{person}{Soohyeon Choi}, \bibinfo{person}{Yong~Kiam Tan}, \bibinfo{person}{Mark~Huasong Meng}, \bibinfo{person}{Mohamed Ragab}, \bibinfo{person}{Soumik Mondal}, \bibinfo{person}{David Mohaisen}, {and} \bibinfo{person}{Khin Mi~Mi Aung}.} \bibinfo{year}{2025}\natexlab{}.
\newblock \bibinfo{title}{I {Can} {Find} {You} in {Seconds}! {Leveraging} {Large} {Language} {Models} for {Code} {Authorship} {Attribution}}.
\newblock
\href{https://doi.org/10.48550/arXiv.2501.08165}{doi:\nolinkurl{10.48550/arXiv.2501.08165}}
\newblock
\shownote{arXiv:2501.08165 [cs] version: 1}.


\bibitem[Cordy and Roy(2011)]%
        {Cordy2011TheNC}
\bibfield{author}{\bibinfo{person}{James~R. Cordy} {and} \bibinfo{person}{Chanchal~Kumar Roy}.} \bibinfo{year}{2011}\natexlab{}.
\newblock \showarticletitle{The NiCad Clone Detector}.
\newblock \bibinfo{journal}{\emph{2011 IEEE 19th International Conference on Program Comprehension}} (\bibinfo{year}{2011}), \bibinfo{pages}{219--220}.
\newblock
\urldef\tempurl%
\url{https://api.semanticscholar.org/CorpusID:2991109}
\showURL{%
\tempurl}


\bibitem[Coskun et~al\mbox{.}(2022)]%
        {coskun_profiling_2022}
\bibfield{author}{\bibinfo{person}{Tugce Coskun}, \bibinfo{person}{Rusen Halepmollasi}, \bibinfo{person}{Khadija Hanifi}, \bibinfo{person}{Ramin~Fadaei Fouladi}, \bibinfo{person}{Pinar~Comak De~Cnudde}, {and} \bibinfo{person}{Ayse Tosun}.} \bibinfo{year}{2022}\natexlab{}.
\newblock \showarticletitle{Profiling developers to predict vulnerable code changes}. In \bibinfo{booktitle}{\emph{Proceedings of the 18th {International} {Conference} on {Predictive} {Models} and {Data} {Analytics} in {Software} {Engineering}}} \emph{(\bibinfo{series}{{PROMISE} 2022})}. \bibinfo{publisher}{Association for Computing Machinery}, \bibinfo{address}{New York, NY, USA}, \bibinfo{pages}{32--41}.
\newblock
\showISBNx{978-1-4503-9860-2}
\href{https://doi.org/10.1145/3558489.3559069}{doi:\nolinkurl{10.1145/3558489.3559069}}


\bibitem[Dakhel et~al\mbox{.}(2023)]%
        {dakhel_dev2vec_2023}
\bibfield{author}{\bibinfo{person}{Arghavan~Moradi Dakhel}, \bibinfo{person}{Michel~C. Desmarais}, {and} \bibinfo{person}{Foutse Khomh}.} \bibinfo{year}{2023}\natexlab{}.
\newblock \showarticletitle{Dev2vec: {Representing} {Domain} {Expertise} of {Developers} in an {Embedding} {Space}}.
\newblock \bibinfo{journal}{\emph{Information and Software Technology}}  \bibinfo{volume}{159} (\bibinfo{date}{July} \bibinfo{year}{2023}), \bibinfo{pages}{107218}.
\newblock
\showISSN{09505849}
\href{https://doi.org/10.1016/j.infsof.2023.107218}{doi:\nolinkurl{10.1016/j.infsof.2023.107218}}
\newblock
\shownote{arXiv:2207.05132 [cs]}.


\bibitem[Dathathri et~al\mbox{.}(2024)]%
        {dathathri_scalable_2024}
\bibfield{author}{\bibinfo{person}{Sumanth Dathathri}, \bibinfo{person}{Abigail See}, \bibinfo{person}{Sumedh Ghaisas}, \bibinfo{person}{Po-Sen Huang}, \bibinfo{person}{Rob McAdam}, \bibinfo{person}{Johannes Welbl}, \bibinfo{person}{Vandana Bachani}, \bibinfo{person}{Alex Kaskasoli}, \bibinfo{person}{Robert Stanforth}, \bibinfo{person}{Tatiana Matejovicova}, \bibinfo{person}{Jamie Hayes}, \bibinfo{person}{Nidhi Vyas}, \bibinfo{person}{Majd~Al Merey}, \bibinfo{person}{Jonah Brown-Cohen}, \bibinfo{person}{Rudy Bunel}, \bibinfo{person}{Borja Balle}, \bibinfo{person}{Taylan Cemgil}, \bibinfo{person}{Zahra Ahmed}, \bibinfo{person}{Kitty Stacpoole}, \bibinfo{person}{Ilia Shumailov}, \bibinfo{person}{Ciprian Baetu}, \bibinfo{person}{Sven Gowal}, \bibinfo{person}{Demis Hassabis}, {and} \bibinfo{person}{Pushmeet Kohli}.} \bibinfo{year}{2024}\natexlab{}.
\newblock \showarticletitle{Scalable watermarking for identifying large language model outputs}.
\newblock \bibinfo{journal}{\emph{Nature}} \bibinfo{volume}{634}, \bibinfo{number}{8035} (\bibinfo{date}{Oct.} \bibinfo{year}{2024}), \bibinfo{pages}{818--823}.
\newblock
\showISSN{1476-4687}
\href{https://doi.org/10.1038/s41586-024-08025-4}{doi:\nolinkurl{10.1038/s41586-024-08025-4}}
\newblock
\shownote{Publisher: Nature Publishing Group}.


\bibitem[Devlin et~al\mbox{.}(2019)]%
        {devlin-etal-2019-bert}
\bibfield{author}{\bibinfo{person}{Jacob Devlin}, \bibinfo{person}{Ming-Wei Chang}, \bibinfo{person}{Kenton Lee}, {and} \bibinfo{person}{Kristina Toutanova}.} \bibinfo{year}{2019}\natexlab{}.
\newblock \showarticletitle{{BERT}: Pre-training of Deep Bidirectional Transformers for Language Understanding}. In \bibinfo{booktitle}{\emph{Proceedings of the 2019 Conference of the North {A}merican Chapter of the Association for Computational Linguistics: Human Language Technologies, Volume 1 (Long and Short Papers)}}, \bibfield{editor}{\bibinfo{person}{Jill Burstein}, \bibinfo{person}{Christy Doran}, {and} \bibinfo{person}{Thamar Solorio}} (Eds.). \bibinfo{publisher}{Association for Computational Linguistics}, \bibinfo{address}{Minneapolis, Minnesota}, \bibinfo{pages}{4171--4186}.
\newblock
\href{https://doi.org/10.18653/v1/N19-1423}{doi:\nolinkurl{10.18653/v1/N19-1423}}


\bibitem[Dirik(2013)]%
        {dirik_source_2013}
\bibfield{author}{\bibinfo{person}{Ahmet~Emir Dirik}.} \bibinfo{year}{2013}\natexlab{}.
\newblock \showarticletitle{Source {Attribution} {Based} on {Physical} {Defects} in {Light} {Path}}.
\newblock In \bibinfo{booktitle}{\emph{Digital {Image} {Forensics}: {There} is {More} to a {Picture} than {Meets} the {Eye}}}, \bibfield{editor}{\bibinfo{person}{Husrev~Taha Sencar} {and} \bibinfo{person}{Nasir Memon}} (Eds.). \bibinfo{publisher}{Springer}, \bibinfo{address}{New York, NY}, \bibinfo{pages}{219--236}.
\newblock
\showISBNx{978-1-4614-0757-7}
\href{https://doi.org/10.1007/978-1-4614-0757-7_7}{doi:\nolinkurl{10.1007/978-1-4614-0757-7_7}}


\bibitem[Feng et~al\mbox{.}(2020)]%
        {feng_codebert_2020}
\bibfield{author}{\bibinfo{person}{Zhangyin Feng}, \bibinfo{person}{Daya Guo}, \bibinfo{person}{Duyu Tang}, \bibinfo{person}{Nan Duan}, \bibinfo{person}{Xiaocheng Feng}, \bibinfo{person}{Ming Gong}, \bibinfo{person}{Linjun Shou}, \bibinfo{person}{Bing Qin}, \bibinfo{person}{Ting Liu}, \bibinfo{person}{Daxin Jiang}, {and} \bibinfo{person}{Ming Zhou}.} \bibinfo{year}{2020}\natexlab{}.
\newblock \bibinfo{title}{{CodeBERT}: {A} {Pre}-{Trained} {Model} for {Programming} and {Natural} {Languages}}.
\newblock
\href{https://doi.org/10.48550/arXiv.2002.08155}{doi:\nolinkurl{10.48550/arXiv.2002.08155}}
\newblock
\shownote{arXiv:2002.08155 [cs]}.


\bibitem[Ferrante et~al\mbox{.}(2016)]%
        {7784595}
\bibfield{author}{\bibinfo{person}{Alberto Ferrante}, \bibinfo{person}{Eric Medvet}, \bibinfo{person}{Francesco Mercaldo}, \bibinfo{person}{Jelena Milosevic}, {and} \bibinfo{person}{Corrado~Aaron Visaggio}.} \bibinfo{year}{2016}\natexlab{}.
\newblock \showarticletitle{Spotting the Malicious Moment: Characterizing Malware Behavior Using Dynamic Features}. In \bibinfo{booktitle}{\emph{2016 11th International Conference on Availability, Reliability and Security (ARES)}}. \bibinfo{pages}{372--381}.
\newblock
\href{https://doi.org/10.1109/ARES.2016.70}{doi:\nolinkurl{10.1109/ARES.2016.70}}


\bibitem[Frantzeskou et~al\mbox{.}(2006a)]%
        {frantzeskou_effective_2006}
\bibfield{author}{\bibinfo{person}{Georgia Frantzeskou}, \bibinfo{person}{Efstathios Stamatatos}, \bibinfo{person}{Stefanos Gritzalis}, {and} \bibinfo{person}{Sokratis Katsikas}.} \bibinfo{year}{2006}\natexlab{a}.
\newblock \showarticletitle{Effective identification of source code authors using byte-level information}. In \bibinfo{booktitle}{\emph{Proceedings of the 28th international conference on {Software} engineering}} \emph{(\bibinfo{series}{{ICSE} '06})}. \bibinfo{publisher}{Association for Computing Machinery}, \bibinfo{address}{New York, NY, USA}, \bibinfo{pages}{893--896}.
\newblock
\showISBNx{978-1-59593-375-1}
\href{https://doi.org/10.1145/1134285.1134445}{doi:\nolinkurl{10.1145/1134285.1134445}}


\bibitem[Frantzeskou et~al\mbox{.}(2006b)]%
        {frantzeskou_source_2006}
\bibfield{author}{\bibinfo{person}{Georgia Frantzeskou}, \bibinfo{person}{Efstathios Stamatatos}, \bibinfo{person}{Stefanos Gritzalis}, {and} \bibinfo{person}{Sokratis Katsikas}.} \bibinfo{year}{2006}\natexlab{b}.
\newblock \showarticletitle{Source {Code} {Author} {Identification} {Based} on {N}-gram {Author} {Profiles}}. In \bibinfo{booktitle}{\emph{Artificial {Intelligence} {Applications} and {Innovations}}}, \bibfield{editor}{\bibinfo{person}{Ilias Maglogiannis}, \bibinfo{person}{Kostas Karpouzis}, {and} \bibinfo{person}{Max Bramer}} (Eds.). \bibinfo{publisher}{Springer US}, \bibinfo{address}{Boston, MA}, \bibinfo{pages}{508--515}.
\newblock
\showISBNx{978-0-387-34224-5}
\href{https://doi.org/10.1007/0-387-34224-9_59}{doi:\nolinkurl{10.1007/0-387-34224-9_59}}


\bibitem[Gray et~al\mbox{.}(2024)]%
        {10.1145/3653973}
\bibfield{author}{\bibinfo{person}{Jason Gray}, \bibinfo{person}{Daniele Sgandurra}, \bibinfo{person}{Lorenzo Cavallaro}, {and} \bibinfo{person}{Jorge Blasco~Alis}.} \bibinfo{year}{2024}\natexlab{}.
\newblock \showarticletitle{Identifying Authorship in Malicious Binaries: Features, Challenges \& Datasets}.
\newblock \bibinfo{journal}{\emph{ACM Comput. Surv.}} \bibinfo{volume}{56}, \bibinfo{number}{8}, Article \bibinfo{articleno}{212} (\bibinfo{date}{April} \bibinfo{year}{2024}), \bibinfo{numpages}{36}~pages.
\newblock
\showISSN{0360-0300}
\href{https://doi.org/10.1145/3653973}{doi:\nolinkurl{10.1145/3653973}}


\bibitem[Guo et~al\mbox{.}(2021)]%
        {guo_graphcodebert_2021}
\bibfield{author}{\bibinfo{person}{Daya Guo}, \bibinfo{person}{Shuo Ren}, \bibinfo{person}{Shuai Lu}, \bibinfo{person}{Zhangyin Feng}, \bibinfo{person}{Duyu Tang}, \bibinfo{person}{Shujie Liu}, \bibinfo{person}{Long Zhou}, \bibinfo{person}{Nan Duan}, \bibinfo{person}{Alexey Svyatkovskiy}, \bibinfo{person}{Shengyu Fu}, \bibinfo{person}{Michele Tufano}, \bibinfo{person}{Shao~Kun Deng}, \bibinfo{person}{Colin Clement}, \bibinfo{person}{Dawn Drain}, \bibinfo{person}{Neel Sundaresan}, \bibinfo{person}{Jian Yin}, \bibinfo{person}{Daxin Jiang}, {and} \bibinfo{person}{Ming Zhou}.} \bibinfo{year}{2021}\natexlab{}.
\newblock \bibinfo{title}{{GraphCodeBERT}: {Pre}-training {Code} {Representations} with {Data} {Flow}}.
\newblock
\href{https://doi.org/10.48550/arXiv.2009.08366}{doi:\nolinkurl{10.48550/arXiv.2009.08366}}
\newblock
\shownote{arXiv:2009.08366 [cs]}.


\bibitem[Guo et~al\mbox{.}(2022)]%
        {guo_method_2022}
\bibfield{author}{\bibinfo{person}{Dixiao Guo}, \bibinfo{person}{Anmin Zhou}, \bibinfo{person}{Liang Liu}, \bibinfo{person}{Shan Liao}, {and} \bibinfo{person}{Lei Zhang}.} \bibinfo{year}{2022}\natexlab{}.
\newblock \showarticletitle{A {Method} of {Source} {Code} {Authorship} {Attribution} {Based} on {Graph} {Neural} {Network}}. In \bibinfo{booktitle}{\emph{Proceedings of 2021 {Chinese} {Intelligent} {Automation} {Conference}}}, \bibfield{editor}{\bibinfo{person}{Zhidong Deng}} (Ed.). \bibinfo{publisher}{Springer}, \bibinfo{address}{Singapore}, \bibinfo{pages}{645--657}.
\newblock
\showISBNx{978-981-16-6372-7}
\href{https://doi.org/10.1007/978-981-16-6372-7_70}{doi:\nolinkurl{10.1007/978-981-16-6372-7_70}}


\bibitem[Guo et~al\mbox{.}(2025)]%
        {guo_codemirage_2025}
\bibfield{author}{\bibinfo{person}{Hanxi Guo}, \bibinfo{person}{Siyuan Cheng}, \bibinfo{person}{Kaiyuan Zhang}, \bibinfo{person}{Guangyu Shen}, {and} \bibinfo{person}{Xiangyu Zhang}.} \bibinfo{year}{2025}\natexlab{}.
\newblock \bibinfo{title}{{CodeMirage}: {A} {Multi}-{Lingual} {Benchmark} for {Detecting} {AI}-{Generated} and {Paraphrased} {Source} {Code} from {Production}-{Level} {LLMs}}.
\newblock
\href{https://doi.org/10.48550/arXiv.2506.11059}{doi:\nolinkurl{10.48550/arXiv.2506.11059}}
\newblock
\shownote{arXiv:2506.11059 [cs]}.


\bibitem[Gupta et~al\mbox{.}(2021)]%
        {gupta_study_2021}
\bibfield{author}{\bibinfo{person}{Surbhi Gupta}, \bibinfo{person}{Neeraj Mohan}, {and} \bibinfo{person}{Munish Kumar}.} \bibinfo{year}{2021}\natexlab{}.
\newblock \showarticletitle{A {Study} on {Source} {Device} {Attribution} {Using} {Still} {Images}}.
\newblock \bibinfo{journal}{\emph{Archives of Computational Methods in Engineering}} \bibinfo{volume}{28}, \bibinfo{number}{4} (\bibinfo{date}{June} \bibinfo{year}{2021}), \bibinfo{pages}{2209--2223}.
\newblock
\showISSN{1886-1784}
\href{https://doi.org/10.1007/s11831-020-09452-y}{doi:\nolinkurl{10.1007/s11831-020-09452-y}}


\bibitem[Gurioli et~al\mbox{.}(2025)]%
        {10992332}
\bibfield{author}{\bibinfo{person}{Andrea Gurioli}, \bibinfo{person}{Maurizio Gabbrielli}, {and} \bibinfo{person}{Stefano Zacchiroli}.} \bibinfo{year}{2025}\natexlab{}.
\newblock \showarticletitle{{ Is This You, LLM? Recognizing AI-written Programs with Multilingual Code Stylometry }}. In \bibinfo{booktitle}{\emph{2025 IEEE International Conference on Software Analysis, Evolution and Reengineering (SANER)}}. \bibinfo{publisher}{IEEE Computer Society}, \bibinfo{address}{Los Alamitos, CA, USA}, \bibinfo{pages}{394--405}.
\newblock
\href{https://doi.org/10.1109/SANER64311.2025.00044}{doi:\nolinkurl{10.1109/SANER64311.2025.00044}}


\bibitem[He et~al\mbox{.}(2023)]%
        {he_debertav3_2023}
\bibfield{author}{\bibinfo{person}{Pengcheng He}, \bibinfo{person}{Jianfeng Gao}, {and} \bibinfo{person}{Weizhu Chen}.} \bibinfo{year}{2023}\natexlab{}.
\newblock \bibinfo{title}{{DeBERTaV3}: {Improving} {DeBERTa} using {ELECTRA}-{Style} {Pre}-{Training} with {Gradient}-{Disentangled} {Embedding} {Sharing}}.
\newblock
\href{https://doi.org/10.48550/arXiv.2111.09543}{doi:\nolinkurl{10.48550/arXiv.2111.09543}}
\newblock
\shownote{arXiv:2111.09543 [cs]}.


\bibitem[He et~al\mbox{.}(2024)]%
        {he_authorship_2024}
\bibfield{author}{\bibinfo{person}{Xie He}, \bibinfo{person}{Arash~Habibi Lashkari}, \bibinfo{person}{Nikhill Vombatkere}, {and} \bibinfo{person}{Dilli~Prasad Sharma}.} \bibinfo{year}{2024}\natexlab{}.
\newblock \showarticletitle{Authorship {Attribution} {Methods}, {Challenges}, and {Future} {Research} {Directions}: {A} {Comprehensive} {Survey}}.
\newblock \bibinfo{journal}{\emph{Information}} \bibinfo{volume}{15}, \bibinfo{number}{3} (\bibinfo{date}{March} \bibinfo{year}{2024}), \bibinfo{pages}{131}.
\newblock
\showISSN{2078-2489}
\href{https://doi.org/10.3390/info15030131}{doi:\nolinkurl{10.3390/info15030131}}
\newblock
\shownote{Number: 3 Publisher: Multidisciplinary Digital Publishing Institute}.


\bibitem[Howard and Ruder(2018)]%
        {howard-ruder-2018-universal}
\bibfield{author}{\bibinfo{person}{Jeremy Howard} {and} \bibinfo{person}{Sebastian Ruder}.} \bibinfo{year}{2018}\natexlab{}.
\newblock \showarticletitle{Universal Language Model Fine-tuning for Text Classification}. In \bibinfo{booktitle}{\emph{Proceedings of the 56th Annual Meeting of the Association for Computational Linguistics (Volume 1: Long Papers)}}, \bibfield{editor}{\bibinfo{person}{Iryna Gurevych} {and} \bibinfo{person}{Yusuke Miyao}} (Eds.). \bibinfo{publisher}{Association for Computational Linguistics}, \bibinfo{address}{Melbourne, Australia}, \bibinfo{pages}{328--339}.
\newblock
\href{https://doi.org/10.18653/v1/P18-1031}{doi:\nolinkurl{10.18653/v1/P18-1031}}


\bibitem[Kalgutkar et~al\mbox{.}(2019)]%
        {kalgutkar_code_2019}
\bibfield{author}{\bibinfo{person}{Vaibhavi Kalgutkar}, \bibinfo{person}{Ratinder Kaur}, \bibinfo{person}{Hugo Gonzalez}, \bibinfo{person}{Natalia Stakhanova}, {and} \bibinfo{person}{Alina Matyukhina}.} \bibinfo{year}{2019}\natexlab{}.
\newblock \showarticletitle{Code {Authorship} {Attribution}: {Methods} and {Challenges}}.
\newblock \bibinfo{journal}{\emph{ACM Comput. Surv.}} \bibinfo{volume}{52}, \bibinfo{number}{1} (\bibinfo{date}{Feb.} \bibinfo{year}{2019}), \bibinfo{pages}{3:1--3:36}.
\newblock
\showISSN{0360-0300}
\href{https://doi.org/10.1145/3292577}{doi:\nolinkurl{10.1145/3292577}}


\bibitem[Khoury et~al\mbox{.}(2023)]%
        {khoury_how_2023}
\bibfield{author}{\bibinfo{person}{Raphaël Khoury}, \bibinfo{person}{Anderson~R. Avila}, \bibinfo{person}{Jacob Brunelle}, {and} \bibinfo{person}{Baba~Mamadou Camara}.} \bibinfo{year}{2023}\natexlab{}.
\newblock \showarticletitle{How {Secure} is {Code} {Generated} by {ChatGPT}?}. In \bibinfo{booktitle}{\emph{2023 {IEEE} {International} {Conference} on {Systems}, {Man}, and {Cybernetics} ({SMC})}}. \bibinfo{pages}{2445--2451}.
\newblock
\href{https://doi.org/10.1109/SMC53992.2023.10394237}{doi:\nolinkurl{10.1109/SMC53992.2023.10394237}}
\newblock
\shownote{ISSN: 2577-1655}.


\bibitem[Kim et~al\mbox{.}(2025)]%
        {kim_marking_2025}
\bibfield{author}{\bibinfo{person}{Jungin Kim}, \bibinfo{person}{Shinwoo Park}, {and} \bibinfo{person}{Yo-Sub Han}.} \bibinfo{year}{2025}\natexlab{}.
\newblock \bibinfo{title}{Marking {Code} {Without} {Breaking} {It}: {Code} {Watermarking} for {Detecting} {LLM}-{Generated} {Code}}.
\newblock
\href{https://doi.org/10.48550/arXiv.2502.18851}{doi:\nolinkurl{10.48550/arXiv.2502.18851}}
\newblock
\shownote{arXiv:2502.18851 [cs]}.


\bibitem[Kumarage et~al\mbox{.}(2024)]%
        {kumarage_survey_2024}
\bibfield{author}{\bibinfo{person}{Tharindu Kumarage}, \bibinfo{person}{Garima Agrawal}, \bibinfo{person}{Paras Sheth}, \bibinfo{person}{Raha Moraffah}, \bibinfo{person}{Aman Chadha}, \bibinfo{person}{Joshua Garland}, {and} \bibinfo{person}{Huan Liu}.} \bibinfo{year}{2024}\natexlab{}.
\newblock \bibinfo{title}{A {Survey} of {AI}-generated {Text} {Forensic} {Systems}: {Detection}, {Attribution}, and {Characterization}}.
\newblock
\href{https://doi.org/10.48550/arXiv.2403.01152}{doi:\nolinkurl{10.48550/arXiv.2403.01152}}
\newblock
\shownote{arXiv:2403.01152 [cs] version: 1}.


\bibitem[Li et~al\mbox{.}(2024)]%
        {li_resilient_2024}
\bibfield{author}{\bibinfo{person}{Boquan Li}, \bibinfo{person}{Mengdi Zhang}, \bibinfo{person}{Peixin Zhang}, \bibinfo{person}{Jun Sun}, {and} \bibinfo{person}{Xingmei Wang}.} \bibinfo{year}{2024}\natexlab{}.
\newblock \bibinfo{title}{Resilient {Watermarking} for {LLM}-{Generated} {Codes}}.
\newblock
\href{https://doi.org/10.48550/arXiv.2402.07518}{doi:\nolinkurl{10.48550/arXiv.2402.07518}}
\newblock
\shownote{arXiv:2402.07518 [cs] version: 1}.


\bibitem[Lissón and Ballier(2018)]%
        {lisson_investigating_2018}
\bibfield{author}{\bibinfo{person}{Paula Lissón} {and} \bibinfo{person}{Nicolas Ballier}.} \bibinfo{year}{2018}\natexlab{}.
\newblock \showarticletitle{Investigating {Lexical} {Progression} through {Lexical} {Diversity} {Metrics} in a {Corpus} of {French} {L3}}.
\newblock \bibinfo{journal}{\emph{Discours. Revue de linguistique, psycholinguistique et informatique. A journal of linguistics, psycholinguistics and computational linguistics}} \bibinfo{number}{23} (\bibinfo{date}{Dec.} \bibinfo{year}{2018}).
\newblock
\showISSN{1963-1723}
\href{https://doi.org/10.4000/discours.9950}{doi:\nolinkurl{10.4000/discours.9950}}
\newblock
\shownote{Number: 23 Publisher: Presses universitaires de Caen}.


\bibitem[Liu et~al\mbox{.}(2023)]%
        {liu_is_2023}
\bibfield{author}{\bibinfo{person}{Jiawei Liu}, \bibinfo{person}{Chunqiu~Steven Xia}, \bibinfo{person}{Yuyao Wang}, {and} \bibinfo{person}{Lingming Zhang}.} \bibinfo{year}{2023}\natexlab{}.
\newblock \showarticletitle{Is {Your} {Code} {Generated} by {ChatGPT} {Really} {Correct}? {Rigorous} {Evaluation} of {Large} {Language} {Models} for {Code} {Generation}}.
\newblock \bibinfo{journal}{\emph{Advances in Neural Information Processing Systems}}  \bibinfo{volume}{36} (\bibinfo{date}{Dec.} \bibinfo{year}{2023}), \bibinfo{pages}{21558--21572}.
\newblock
\urldef\tempurl%
\url{https://proceedings.neurips.cc/paper_files/paper/2023/hash/43e9d647ccd3e4b7b5baab53f0368686-Abstract-Conference.html}
\showURL{%
\tempurl}


\bibitem[Liu et~al\mbox{.}(2019)]%
        {liu_roberta_2019}
\bibfield{author}{\bibinfo{person}{Yinhan Liu}, \bibinfo{person}{Myle Ott}, \bibinfo{person}{Naman Goyal}, \bibinfo{person}{Jingfei Du}, \bibinfo{person}{Mandar Joshi}, \bibinfo{person}{Danqi Chen}, \bibinfo{person}{Omer Levy}, \bibinfo{person}{Mike Lewis}, \bibinfo{person}{Luke Zettlemoyer}, {and} \bibinfo{person}{Veselin Stoyanov}.} \bibinfo{year}{2019}\natexlab{}.
\newblock \bibinfo{title}{{RoBERTa}: {A} {Robustly} {Optimized} {BERT} {Pretraining} {Approach}}.
\newblock
\href{https://doi.org/10.48550/arXiv.1907.11692}{doi:\nolinkurl{10.48550/arXiv.1907.11692}}
\newblock
\shownote{arXiv:1907.11692 [cs]}.


\bibitem[McCabe(1976)]%
        {McCabe}
\bibfield{author}{\bibinfo{person}{T.J. McCabe}.} \bibinfo{year}{1976}\natexlab{}.
\newblock \showarticletitle{A Complexity Measure}.
\newblock \bibinfo{journal}{\emph{IEEE Transactions on Software Engineering}} \bibinfo{volume}{SE-2}, \bibinfo{number}{4} (\bibinfo{year}{1976}), \bibinfo{pages}{308--320}.
\newblock
\href{https://doi.org/10.1109/TSE.1976.233837}{doi:\nolinkurl{10.1109/TSE.1976.233837}}


\bibitem[Mitchell et~al\mbox{.}(2023)]%
        {mitchell_detectgpt_2023}
\bibfield{author}{\bibinfo{person}{Eric Mitchell}, \bibinfo{person}{Yoonho Lee}, \bibinfo{person}{Alexander Khazatsky}, \bibinfo{person}{Christopher~D. Manning}, {and} \bibinfo{person}{Chelsea Finn}.} \bibinfo{year}{2023}\natexlab{}.
\newblock \bibinfo{title}{{DetectGPT}: {Zero}-{Shot} {Machine}-{Generated} {Text} {Detection} using {Probability} {Curvature}}.
\newblock
\href{https://doi.org/10.48550/arXiv.2301.11305}{doi:\nolinkurl{10.48550/arXiv.2301.11305}}
\newblock
\shownote{arXiv:2301.11305 [cs]}.


\bibitem[Nguyen et~al\mbox{.}(2024)]%
        {nguyen_gptsniffer_2024}
\bibfield{author}{\bibinfo{person}{Phuong~T. Nguyen}, \bibinfo{person}{Juri Di~Rocco}, \bibinfo{person}{Claudio Di~Sipio}, \bibinfo{person}{Riccardo Rubei}, \bibinfo{person}{Davide Di~Ruscio}, {and} \bibinfo{person}{Massimiliano Di~Penta}.} \bibinfo{year}{2024}\natexlab{}.
\newblock \showarticletitle{{GPTSniffer}: {A} {CodeBERT}-based classifier to detect source code written by {ChatGPT}}.
\newblock \bibinfo{journal}{\emph{Journal of Systems and Software}}  \bibinfo{volume}{214} (\bibinfo{date}{Aug.} \bibinfo{year}{2024}), \bibinfo{pages}{112059}.
\newblock
\showISSN{0164-1212}
\href{https://doi.org/10.1016/j.jss.2024.112059}{doi:\nolinkurl{10.1016/j.jss.2024.112059}}


\bibitem[Paek and Mohan(2025)]%
        {paek_detection_2025}
\bibfield{author}{\bibinfo{person}{Timothy Paek} {and} \bibinfo{person}{Chilukuri Mohan}.} \bibinfo{year}{2025}\natexlab{}.
\newblock \bibinfo{title}{Detection of {LLM}-{Generated} {Java} {Code} {Using} {Discretized} {Nested} {Bigrams}}.
\newblock
\href{https://doi.org/10.48550/arXiv.2502.15740}{doi:\nolinkurl{10.48550/arXiv.2502.15740}}
\newblock
\shownote{arXiv:2502.15740 [cs]}.


\bibitem[Pan et~al\mbox{.}(2024)]%
        {pan_assessing_2024}
\bibfield{author}{\bibinfo{person}{Wei~Hung Pan}, \bibinfo{person}{Ming~Jie Chok}, \bibinfo{person}{Jonathan Leong~Shan Wong}, \bibinfo{person}{Yung~Xin Shin}, \bibinfo{person}{Yeong~Shian Poon}, \bibinfo{person}{Zhou Yang}, \bibinfo{person}{Chun~Yong Chong}, \bibinfo{person}{David Lo}, {and} \bibinfo{person}{Mei~Kuan Lim}.} \bibinfo{year}{2024}\natexlab{}.
\newblock \bibinfo{title}{Assessing {AI} {Detectors} in {Identifying} {AI}-{Generated} {Code}: {Implications} for {Education}}.
\newblock
\href{https://doi.org/10.48550/arXiv.2401.03676}{doi:\nolinkurl{10.48550/arXiv.2401.03676}}
\newblock
\shownote{arXiv:2401.03676 [cs]}.


\bibitem[Park et~al\mbox{.}(2025)]%
        {park_detection_2025}
\bibfield{author}{\bibinfo{person}{Shinwoo Park}, \bibinfo{person}{Hyundong Jin}, \bibinfo{person}{Jeong-won Cha}, {and} \bibinfo{person}{Yo-Sub Han}.} \bibinfo{year}{2025}\natexlab{}.
\newblock \bibinfo{title}{Detection of {LLM}-{Paraphrased} {Code} and {Identification} of the {Responsible} {LLM} {Using} {Coding} {Style} {Features}}.
\newblock
\href{https://doi.org/10.48550/arXiv.2502.17749}{doi:\nolinkurl{10.48550/arXiv.2502.17749}}
\newblock
\shownote{arXiv:2502.17749 [cs] version: 2}.


\bibitem[Quiring et~al\mbox{.}(2019)]%
        {quiring_misleading_2019}
\bibfield{author}{\bibinfo{person}{Erwin Quiring}, \bibinfo{person}{Alwin Maier}, {and} \bibinfo{person}{Konrad Rieck}.} \bibinfo{year}{2019}\natexlab{}.
\newblock \showarticletitle{Misleading {Authorship} {Attribution} of {Source} {Code} using {Adversarial} {Learning}}. \bibinfo{pages}{479--496}.
\newblock
\showISBNx{978-1-939133-06-9}
\urldef\tempurl%
\url{https://www.usenix.org/conference/usenixsecurity19/presentation/quiring}
\showURL{%
\tempurl}


\bibitem[Rosenblum et~al\mbox{.}(2011)]%
        {rosenblum_who_2011}
\bibfield{author}{\bibinfo{person}{Nathan Rosenblum}, \bibinfo{person}{Xiaojin Zhu}, {and} \bibinfo{person}{Barton~P. Miller}.} \bibinfo{year}{2011}\natexlab{}.
\newblock \showarticletitle{Who {Wrote} {This} {Code}? {Identifying} the {Authors} of {Program} {Binaries}}. In \bibinfo{booktitle}{\emph{Computer {Security} – {ESORICS} 2011}}, \bibfield{editor}{\bibinfo{person}{Vijay Atluri} {and} \bibinfo{person}{Claudia Diaz}} (Eds.). \bibinfo{publisher}{Springer}, \bibinfo{address}{Berlin, Heidelberg}, \bibinfo{pages}{172--189}.
\newblock
\showISBNx{978-3-642-23822-2}
\href{https://doi.org/10.1007/978-3-642-23822-2_10}{doi:\nolinkurl{10.1007/978-3-642-23822-2_10}}


\bibitem[Sanh et~al\mbox{.}(2020)]%
        {sanh_distilbert_2020}
\bibfield{author}{\bibinfo{person}{Victor Sanh}, \bibinfo{person}{Lysandre Debut}, \bibinfo{person}{Julien Chaumond}, {and} \bibinfo{person}{Thomas Wolf}.} \bibinfo{year}{2020}\natexlab{}.
\newblock \bibinfo{title}{{DistilBERT}, a distilled version of {BERT}: smaller, faster, cheaper and lighter}.
\newblock
\href{https://doi.org/10.48550/arXiv.1910.01108}{doi:\nolinkurl{10.48550/arXiv.1910.01108}}
\newblock
\shownote{arXiv:1910.01108 [cs]}.


\bibitem[Seroussi et~al\mbox{.}(2014)]%
        {seroussi_authorship_2014}
\bibfield{author}{\bibinfo{person}{Yanir Seroussi}, \bibinfo{person}{Ingrid Zukerman}, {and} \bibinfo{person}{Fabian Bohnert}.} \bibinfo{year}{2014}\natexlab{}.
\newblock \showarticletitle{Authorship {Attribution} with {Topic} {Models}}.
\newblock \bibinfo{journal}{\emph{Computational Linguistics}} \bibinfo{volume}{40}, \bibinfo{number}{2} (\bibinfo{date}{June} \bibinfo{year}{2014}), \bibinfo{pages}{269--310}.
\newblock
\href{https://doi.org/10.1162/COLI_a_00173}{doi:\nolinkurl{10.1162/COLI_a_00173}}
\newblock
\shownote{Place: Cambridge, MA Publisher: MIT Press}.


\bibitem[Silva et~al\mbox{.}(2024)]%
        {silva_forged-gan-bert_2024}
\bibfield{author}{\bibinfo{person}{Kanishka Silva}, \bibinfo{person}{Ingo Frommholz}, \bibinfo{person}{Burcu Can}, \bibinfo{person}{Fred Blain}, \bibinfo{person}{Raheem Sarwar}, {and} \bibinfo{person}{Laura Ugolini}.} \bibinfo{year}{2024}\natexlab{}.
\newblock \showarticletitle{Forged-{GAN}-{BERT}: {Authorship} {Attribution} for {LLM}-{Generated} {Forged} {Novels}}. In \bibinfo{booktitle}{\emph{Proceedings of the 18th {Conference} of the {European} {Chapter} of the {Association} for {Computational} {Linguistics}: {Student} {Research} {Workshop}}}, \bibfield{editor}{\bibinfo{person}{Neele Falk}, \bibinfo{person}{Sara Papi}, {and} \bibinfo{person}{Mike Zhang}} (Eds.). \bibinfo{publisher}{Association for Computational Linguistics}, \bibinfo{address}{St. Julian's, Malta}, \bibinfo{pages}{325--337}.
\newblock
\urldef\tempurl%
\url{https://aclanthology.org/2024.eacl-srw.26/}
\showURL{%
\tempurl}


\bibitem[Song et~al\mbox{.}(2022a)]%
        {9825799}
\bibfield{author}{\bibinfo{person}{Qige Song}, \bibinfo{person}{Yongzheng Zhang}, \bibinfo{person}{Linshu Ouyang}, {and} \bibinfo{person}{Yige Chen}.} \bibinfo{year}{2022}\natexlab{a}.
\newblock \showarticletitle{{ BinMLM: Binary Authorship Verification with Flow-aware Mixture-of-Shared Language Model }}. In \bibinfo{booktitle}{\emph{2022 IEEE International Conference on Software Analysis, Evolution and Reengineering (SANER)}}. \bibinfo{publisher}{IEEE Computer Society}, \bibinfo{address}{Los Alamitos, CA, USA}, \bibinfo{pages}{1023--1033}.
\newblock
\showISSN{1534-5351}
\href{https://doi.org/10.1109/SANER53432.2022.00120}{doi:\nolinkurl{10.1109/SANER53432.2022.00120}}


\bibitem[Song et~al\mbox{.}(2022b)]%
        {song_binmlm_2022}
\bibfield{author}{\bibinfo{person}{Qige Song}, \bibinfo{person}{Yongzheng Zhang}, \bibinfo{person}{Linshu Ouyang}, {and} \bibinfo{person}{Yige Chen}.} \bibinfo{year}{2022}\natexlab{b}.
\newblock \bibinfo{title}{{BinMLM}: {Binary} {Authorship} {Verification} with {Flow}-aware {Mixture}-of-{Shared} {Language} {Model}}.
\newblock
\href{https://doi.org/10.48550/arXiv.2203.04472}{doi:\nolinkurl{10.48550/arXiv.2203.04472}}
\newblock
\shownote{arXiv:2203.04472 [cs]}.


\bibitem[Suh et~al\mbox{.}(2024)]%
        {suh_empirical_2024}
\bibfield{author}{\bibinfo{person}{Hyunjae Suh}, \bibinfo{person}{Mahan Tafreshipour}, \bibinfo{person}{Jiawei Li}, \bibinfo{person}{Adithya Bhattiprolu}, {and} \bibinfo{person}{Iftekhar Ahmed}.} \bibinfo{year}{2024}\natexlab{}.
\newblock \bibinfo{title}{An {Empirical} {Study} on {Automatically} {Detecting} {AI}-{Generated} {Source} {Code}: {How} {Far} {Are} {We}?}
\newblock
\href{https://doi.org/10.48550/arXiv.2411.04299}{doi:\nolinkurl{10.48550/arXiv.2411.04299}}
\newblock
\shownote{arXiv:2411.04299 [cs] version: 1}.


\bibitem[Suresh et~al\mbox{.}(2025)]%
        {suresh_is_2025}
\bibfield{author}{\bibinfo{person}{Tarun Suresh}, \bibinfo{person}{Shubham Ugare}, \bibinfo{person}{Gagandeep Singh}, {and} \bibinfo{person}{Sasa Misailovic}.} \bibinfo{year}{2025}\natexlab{}.
\newblock \bibinfo{title}{Is {The} {Watermarking} {Of} {LLM}-{Generated} {Code} {Robust}?}
\newblock
\href{https://doi.org/10.48550/arXiv.2403.17983}{doi:\nolinkurl{10.48550/arXiv.2403.17983}}
\newblock
\shownote{arXiv:2403.17983 [cs]}.


\bibitem[Thathsarani(2024)]%
        {thathsarani_comprehensive_2024}
\bibfield{author}{\bibinfo{person}{N.~N. Thathsarani}.} \bibinfo{year}{2024}\natexlab{}.
\newblock \showarticletitle{A {Comprehensive} {Software} {Complexity} {Metric} {Based} on {Cyclomatic} {Complexity}}. In \bibinfo{booktitle}{\emph{2024 4th {International} {Conference} of {Science} and {Information} {Technology} in {Smart} {Administration} ({ICSINTESA})}}. \bibinfo{pages}{90--95}.
\newblock
\href{https://doi.org/10.1109/ICSINTESA62455.2024.10748227}{doi:\nolinkurl{10.1109/ICSINTESA62455.2024.10748227}}


\bibitem[Tihanyi et~al\mbox{.}(2024)]%
        {tihanyi_how_2024}
\bibfield{author}{\bibinfo{person}{Norbert Tihanyi}, \bibinfo{person}{Tamas Bisztray}, \bibinfo{person}{Mohamed~Amine Ferrag}, \bibinfo{person}{Ridhi Jain}, {and} \bibinfo{person}{Lucas~C. Cordeiro}.} \bibinfo{year}{2024}\natexlab{}.
\newblock \showarticletitle{How secure is {AI}-generated code: a large-scale comparison of large language models}.
\newblock \bibinfo{journal}{\emph{Empirical Software Engineering}} \bibinfo{volume}{30}, \bibinfo{number}{2} (\bibinfo{date}{Dec.} \bibinfo{year}{2024}), \bibinfo{pages}{47}.
\newblock
\showISSN{1573-7616}
\href{https://doi.org/10.1007/s10664-024-10590-1}{doi:\nolinkurl{10.1007/s10664-024-10590-1}}


\bibitem[Ullah et~al\mbox{.}(2022)]%
        {ullah_crolssim_2022}
\bibfield{author}{\bibinfo{person}{Farhan Ullah}, \bibinfo{person}{Muhammad~Rashid Naeem}, \bibinfo{person}{Hamad Naeem}, \bibinfo{person}{Xiaochun Cheng}, {and} \bibinfo{person}{Mamoun Alazab}.} \bibinfo{year}{2022}\natexlab{}.
\newblock \showarticletitle{{CroLSSim}: {Cross}-language software similarity detector using hybrid approach of {LSA}-based {AST}-{MDrep} features and {CNN}-{LSTM} model}.
\newblock \bibinfo{journal}{\emph{International Journal of Intelligent Systems}} \bibinfo{volume}{37}, \bibinfo{number}{9} (\bibinfo{year}{2022}), \bibinfo{pages}{5768--5795}.
\newblock
\showISSN{1098-111X}
\href{https://doi.org/10.1002/int.22813}{doi:\nolinkurl{10.1002/int.22813}}
\newblock
\shownote{\_eprint: https://onlinelibrary.wiley.com/doi/pdf/10.1002/int.22813}.


\bibitem[Vaswani et~al\mbox{.}(2017)]%
        {vaswani_attention_2017}
\bibfield{author}{\bibinfo{person}{Ashish Vaswani}, \bibinfo{person}{Noam Shazeer}, \bibinfo{person}{Niki Parmar}, \bibinfo{person}{Jakob Uszkoreit}, \bibinfo{person}{Llion Jones}, \bibinfo{person}{Aidan~N Gomez}, \bibinfo{person}{Ł~ukasz Kaiser}, {and} \bibinfo{person}{Illia Polosukhin}.} \bibinfo{year}{2017}\natexlab{}.
\newblock \showarticletitle{Attention is {All} you {Need}}. In \bibinfo{booktitle}{\emph{Advances in {Neural} {Information} {Processing} {Systems}}}, Vol.~\bibinfo{volume}{30}. \bibinfo{publisher}{Curran Associates, Inc.}
\newblock
\urldef\tempurl%
\url{https://papers.nips.cc/paper_files/paper/2017/hash/3f5ee243547dee91fbd053c1c4a845aa-Abstract.html}
\showURL{%
\tempurl}


\bibitem[Wang et~al\mbox{.}(2018)]%
        {wang_integration_2018}
\bibfield{author}{\bibinfo{person}{Ningfei Wang}, \bibinfo{person}{Shouling Ji}, {and} \bibinfo{person}{Ting Wang}.} \bibinfo{year}{2018}\natexlab{}.
\newblock \showarticletitle{Integration of {Static} and {Dynamic} {Code} {Stylometry} {Analysis} for {Programmer} {De}-anonymization}. In \bibinfo{booktitle}{\emph{Proceedings of the 11th {ACM} {Workshop} on {Artificial} {Intelligence} and {Security}}} \emph{(\bibinfo{series}{{AISec} '18})}. \bibinfo{publisher}{Association for Computing Machinery}, \bibinfo{address}{New York, NY, USA}, \bibinfo{pages}{74--84}.
\newblock
\showISBNx{978-1-4503-6004-3}
\href{https://doi.org/10.1145/3270101.3270110}{doi:\nolinkurl{10.1145/3270101.3270110}}


\bibitem[Wang et~al\mbox{.}(2021)]%
        {wang_codet5_2021}
\bibfield{author}{\bibinfo{person}{Yue Wang}, \bibinfo{person}{Weishi Wang}, \bibinfo{person}{Shafiq Joty}, {and} \bibinfo{person}{Steven C.~H. Hoi}.} \bibinfo{year}{2021}\natexlab{}.
\newblock \bibinfo{title}{{CodeT5}: {Identifier}-aware {Unified} {Pre}-trained {Encoder}-{Decoder} {Models} for {Code} {Understanding} and {Generation}}.
\newblock
\href{https://doi.org/10.48550/arXiv.2109.00859}{doi:\nolinkurl{10.48550/arXiv.2109.00859}}
\newblock
\shownote{arXiv:2109.00859 [cs]}.


\bibitem[Warner et~al\mbox{.}(2024)]%
        {modernBERT}
\bibfield{author}{\bibinfo{person}{Benjamin Warner}, \bibinfo{person}{Antoine Chaffin}, \bibinfo{person}{Benjamin Clavié}, \bibinfo{person}{Orion Weller}, \bibinfo{person}{Oskar Hallström}, \bibinfo{person}{Said Taghadouini}, \bibinfo{person}{Alexis Gallagher}, \bibinfo{person}{Raja Biswas}, \bibinfo{person}{Faisal Ladhak}, \bibinfo{person}{Tom Aarsen}, \bibinfo{person}{Nathan Cooper}, \bibinfo{person}{Griffin Adams}, \bibinfo{person}{Jeremy Howard}, {and} \bibinfo{person}{Iacopo Poli}.} \bibinfo{year}{2024}\natexlab{}.
\newblock \bibinfo{title}{Smarter, Better, Faster, Longer: A Modern Bidirectional Encoder for Fast, Memory Efficient, and Long Context Finetuning and Inference}.
\newblock
\showeprint[arxiv]{2412.13663}~[cs.CL]
\urldef\tempurl%
\url{https://arxiv.org/abs/2412.13663}
\showURL{%
\tempurl}


\bibitem[Wei et~al\mbox{.}(2022)]%
        {wei_chain--thought_2022}
\bibfield{author}{\bibinfo{person}{Jason Wei}, \bibinfo{person}{Xuezhi Wang}, \bibinfo{person}{Dale Schuurmans}, \bibinfo{person}{Maarten Bosma}, \bibinfo{person}{Brian Ichter}, \bibinfo{person}{Fei Xia}, \bibinfo{person}{Ed Chi}, \bibinfo{person}{Quoc Le}, {and} \bibinfo{person}{Denny Zhou}.} \bibinfo{year}{2022}\natexlab{}.
\newblock \bibinfo{title}{Chain-of-{Thought} {Prompting} {Elicits} {Reasoning} in {Large} {Language} {Models}}.
\newblock
\urldef\tempurl%
\url{https://arxiv.org/abs/2201.11903v6}
\showURL{%
\tempurl}


\bibitem[Xu and Sheng(2025)]%
        {xu_codevision_2025}
\bibfield{author}{\bibinfo{person}{Zhenyu Xu} {and} \bibinfo{person}{Victor~S. Sheng}.} \bibinfo{year}{2025}\natexlab{}.
\newblock \bibinfo{title}{{CodeVision}: {Detecting} {LLM}-{Generated} {Code} {Using} {2D} {Token} {Probability} {Maps} and {Vision} {Models}}.
\newblock
\href{https://doi.org/10.48550/arXiv.2501.03288}{doi:\nolinkurl{10.48550/arXiv.2501.03288}}
\newblock
\shownote{arXiv:2501.03288 [cs] version: 1}.


\bibitem[Zafar et~al\mbox{.}(2020)]%
        {zafar_language_2020}
\bibfield{author}{\bibinfo{person}{Sarim Zafar}, \bibinfo{person}{Muhammad~Usman Sarwar}, \bibinfo{person}{Saeed Salem}, {and} \bibinfo{person}{Muhammad~Zubair Malik}.} \bibinfo{year}{2020}\natexlab{}.
\newblock \showarticletitle{Language and {Obfuscation} {Oblivious} {Source} {Code} {Authorship} {Attribution}}.
\newblock \bibinfo{journal}{\emph{IEEE Access}}  \bibinfo{volume}{8} (\bibinfo{year}{2020}), \bibinfo{pages}{197581--197596}.
\newblock
\showISSN{2169-3536}
\href{https://doi.org/10.1109/ACCESS.2020.3034932}{doi:\nolinkurl{10.1109/ACCESS.2020.3034932}}


\bibitem[Álvarez Fidalgo and Ortin(2025)]%
        {alvarez-fidalgo_clave_2025}
\bibfield{author}{\bibinfo{person}{David Álvarez Fidalgo} {and} \bibinfo{person}{Francisco Ortin}.} \bibinfo{year}{2025}\natexlab{}.
\newblock \showarticletitle{{CLAVE}: {A} deep learning model for source code authorship verification with contrastive learning and transformer encoders}.
\newblock \bibinfo{journal}{\emph{Inf. Process. Manage.}} \bibinfo{volume}{62}, \bibinfo{number}{3} (\bibinfo{date}{April} \bibinfo{year}{2025}).
\newblock
\showISSN{0306-4573}
\href{https://doi.org/10.1016/j.ipm.2024.104005}{doi:\nolinkurl{10.1016/j.ipm.2024.104005}}


\end{thebibliography}

\end{document}